\documentclass[mathpazo]{cicp}
\usepackage{subfigure}
\usepackage{algorithmic}
\usepackage{amsmath}
\usepackage{amssymb}
\usepackage{multirow}
\usepackage{makecell}
\usepackage{graphicx}
\usepackage{xcolor}

\usepackage{url}
\begin{document}
%%%%% title : short title may not be used but TITLE is required.
% \title{TITLE}
% \title[short title]{TITLE}
\title{Curriculum Learning of Physics-Informed Neural Networks based on Spatial Correlation}

%%%%% author(s) :
% single author:
% \author[name in running head]{AUTHOR\corrauth}
% [name in running head] is NOT OPTIONAL, it is a MUST.
% Use \corrauth to indicate the corresponding author.
% Use \email to provide email address of author.
% \footnote and \thanks are not used in the heading section.
% Another acknowlegments/support of grants, state in Acknowledgments section
% \section*{Acknowledgments}

% \author[O.~Author]{Xujia Chen\corrauth, Daming Shi, Yi Liu and Wenhui Fan}
% \address{Department of Automation, Tsinghua University,
% Beijing 100084, P.R. China}
% \email{{\tt chenxj20@mails.tsinghua.edu.cn} (O.~Author)}

% \author[O.~Author]{Authors}
% \address{Address}
% \email{{\tt emails} (O.~Author)}

\author[Chen X J et.~al.]{Xujia Chen\corrauth, Xinyue Hu, Letian Chen, Daming Shi and Wenhui Fan}
\address{Department of Automation, Tsinghua University,
Beijing 100084, P.R. China}
\emails{{\tt chenxj20@mails.tsinghua.edu.cn} (X.~Chen)
%   {\tt hu-xy25@mails.tsinghua.edu.cn} (X.~Hu),
%   {\tt clt21@mails.tsinghua.edu.cn} (L.~Chen),
%   {\tt shidm18@tsinghua.org.cn} (D.~Shi),
% {\tt fanwenhui@tsinghua.edu.cn} (W.~Fan)
}

% multiple authors:
% Note the use of \affil and \affilnum to link names and addresses.
% The author for correspondence is marked by \corrauth.
% use \emails to provide email addresses of authors
% e.g. below example has 3 authors, first author is also the corresponding
%      author, author 1 and 3 having the same address.
% \author[Zhang Z R et.~al.]{Zhengru Zhang\affil{1}\comma\corrauth,
%       Author Chan\affil{2}, and Author Zhao\affil{1}}
% \address{\affilnum{1}\ School of Mathematical Sciences,
%          Beijing Normal University,
%          Beijing 100875, P.R. China. \\
%           \affilnum{2}\ Department of Mathematics,
%           Hong Kong Baptist University, Hong Kong SAR}
% \emails{{\tt zhang@email} (Z.~Zhang), {\tt chan@email} (A.~Chan),
%          {\tt zhao@email} (A.~Zhao)}
% \footnote and \thanks are not used in the heading section.
% Another acknowlegments/support of grants, state in Acknowledgments section
% \section*{Acknowledgments}

%%%%% Begin Abstract %%%%%%%%%%%
\begin{abstract}
  Physics-Informed Neural Networks (PINNs) combine deep learning with physical constraints for solving partial differential equations (PDEs), and are widely applied in fluid mechanics, heat transfer, and solid mechanics.
  However, PINN training still suffers from high-dimensional non-convex loss landscapes, imbalanced multi-objective constraints, and ineffective information propagation.
  Existing curriculum learning and causality-guided strategies improve training stability, but mainly focus on temporal or parametric progression, lacking explicit treatment of spatial information propagation and inter-region consistency.
  Moreover, they are not directly applicable to boundary value problems (BVPs) with strong spatial coupling.
  To address this issue, we propose a spatially correlated curriculum learning framework for PINNs.
  To the best of our knowledge, this is the first work to address PINN training difficulties from the perspective of spatial coupling among subregions.
  First, spatial causal weights guide information from near-boundary regions inward, reducing optimization failures and spurious convergence.
  Second, a low-frequency information bridge enforces pseudo-label-based consistency across spatially separated regions, suppressing global low-frequency drift.
  Third, a region-adaptive reweighting strategy adjusts subregion losses to reduce local residuals and recover high-frequency details.
  Experiments on PDE benchmarks show that, under comparable computational cost, the proposed method alleviates training failures and improves solution accuracy.
  The code is available at \url{https://github.com/pigofmomo/CurriculumLearningPINN}.
\end{abstract}
%%%%% end %%%%%%%%%%%

%%%%% AMS/PACs/Keywords %%%%%%%%%%%
%\pac{}

%%%% maketitle %%%%%
\maketitle

%%%% Start %%%%%%
\section{Introduction}
\label{sec:intro}

The evolution of many physical systems in nature is typically described by partial differential equations (PDEs). However, in practical engineering and scientific computations, PDEs often exhibit characteristics such as strong nonlinearity, multi-scale phenomena, and complex geometries, making the efficient and accurate solution of these equations one of the central challenges in simulations. 
Traditional numerical methods (such as finite differences, finite elements, and finite volumes) typically require fine grid discretization and complex preprocessing steps when solving PDEs\textsuperscript{\cite{ames2014numerical}}. This becomes especially problematic when dealing with high-dimensional problems, performing real-time predictions, or solving inverse problems, where traditional methods often encounter severe computational and storage bottlenecks, leading to inefficiency.

In recent years, AI-based PDE solving research has developed rapidly\textsuperscript{\cite{brunton2024promising}}. Physics-Informed Neural Networks (PINNs)\textsuperscript{\cite{raissi2019physics}}, as a PDE solving method that integrates deep learning with physical constraints, enable end-to-end fitting of the solution function. This approach avoids complex procedures such as explicit grid generation, and has the potential to simultaneously address forward and inverse problems, perform parameter identification, and facilitate scientific discovery\textsuperscript{\cite{zhang2026physics}}. As a result, PINNs have gained widespread attention and application in engineering fields such as fluid mechanics, heat transfer, solid mechanics, and electromagnetics\textsuperscript{\cite{fan2026embedding}}. 

The basic idea of PINNs is to use neural networks as a parametric representation of continuous functions. Specifically, PINNs typically use architectures such as multi-layer perceptrons (MLP) as the backbone network, mapping spatial-temporal coordinates and physical parameters to state variables. During the training phase, the PDE residual term is explicitly incorporated into the loss function, and the solution's partial derivatives with respect to the inputs are computed using automatic differentiation, thus transforming the constraints imposed by the differential operators into an optimizable cost function. Subsequently, first- or second-order optimizers (such as Adam, L-BFGS, etc.) are employed to perform gradient descent on the network weights, enabling the network output to simultaneously satisfy the equation constraints and initial/boundary conditions.

Despite the simplicity of this framework, its training process often faces significant difficulties\textsuperscript{\cite{krishnapriyan2021characterizing, rathore2024challenges, wang2022and}}: in the absence of sufficient supervised data, the PDE operator tends to dominate the optimization direction of the solution within the domain. 
The PDE operator terms often exhibit "stiffness" characteristics, such as strong nonlinearity, multi-scale variations, and potential instability, leading to slow loss function convergence. 
At the same time, conflicts can arise between the PDE residual, boundary/initial value constraints, and possibly data terms, making PINN training inherently a multi-objective optimization problem, further exacerbating optimization difficulties. 
Additionally, the overall loss landscape is highly non-convex, causing the model to easily get trapped in poor local minima or plateau regions, resulting in a long-term failure of error convergence.

To address the aforementioned training failure issues, various improvement strategies have been proposed in recent years, with the core goal of reducing optimization difficulties and improving the efficiency and accuracy of PINN solutions. 
One such strategy is curriculum learning\textsuperscript{\cite{wang2021survey}}, which first trains the model on relatively simple tasks, such as smooth solutions, weak nonlinearity, or localized regions, and then gradually increases the difficulty by introducing high-frequency components, stronger advection/reaction effects, more complex boundary conditions, or denser sampling. 
This progressive training process provides a clearer optimization path and reduces the likelihood of convergence to poor local minima.
Another important approach is causality-guided training\textsuperscript{\cite{WANG2024116813}}: for PDEs with explicit temporal causality (such as transient problems where information propagates over time), sequential constraints or staged progression are applied in the time dimension, making the model prioritize fitting the solution at earlier time steps and gradually extend to later time steps, avoiding "incorrect propagation" caused by violating causality.

In addition, existing studies have improved PINNs from multiple perspectives. Weight adjustment dynamically balances different loss terms to mitigate multi-objective conflicts\textsuperscript{\cite{wang2022and}}, adaptive sampling refines collocation points in high-residual regions to better capture local details\textsuperscript{\cite{daw2022mitigating}}, and improved optimizers and training strategies enhance optimization stability and convergence\textsuperscript{\cite{liu2025config}}. Structural priors further strengthen the representation of multi-scale features\textsuperscript{\cite{huang2025frequency}}. Moreover, domain decomposition splits a complex PDE into local subproblems for collaborative learning\textsuperscript{\cite{dolean2024multilevel}}; related efforts also combine temporal decomposition with numerical solvers to accelerate subdomain convergence\textsuperscript{\cite{meng2020ppinn}}, incorporate finite-volume-style local conservation residuals to improve fine-scale accuracy\textsuperscript{\cite{yang2025s}}, and introduce extra correlation or penalty terms to better coordinate competing objectives\textsuperscript{\cite{cao2025wbpinn}}.

However, existing PINN improvement strategies, including curriculum learning and causality-guided training, still suffer from notable limitations. 
First, time-causality-guided methods rely on a unidirectional propagation structure along the temporal dimension. For many coupled PDE boundary value problems (BVPs), however, no such one-way causality exists; instead, the solution is primarily governed by spatial coupling and boundary constraints. As a result, time-stepping-based causal training paradigms are not directly applicable. Second, existing curriculum learning methods usually organize training according to task difficulty, such as from low to high frequency, from coarse to fine, or from local to global. Yet they generally do not explicitly account for spatial heterogeneity of the solution, inter-region dependency, or information propagation pathways across the domain. Consequently, under strongly coupled spatial settings, they may still yield locally converged solutions that remain globally inconsistent.

To address these limitations, this paper proposes a spatially correlated curriculum framework for PINNs, which guides the training process by explicitly considering spatial information propagation and inter-region consistency. Specifically:

\begin{itemize}
    \item We propose a spatially correlated curriculum learning framework that uses spatial causal weights to guide solution information from near-boundary regions toward the interior, promoting effective spatial information propagation while suppressing erroneous propagation.

    \item We design an information bridge mechanism to enhance communication between mutually coupled or long-range correlated regions, encouraging consistent solution patterns in intermediate areas and reducing cross-region low-frequency drift.

    \item We introduce a region-adaptive reweighting strategy based on regional PDE residuals and gradient contributions, which reduces local residuals and improves the recovery of high-frequency details.
\end{itemize}

Extensive experiments demonstrate that the proposed framework improves both global consistency and local detail recovery, providing an effective new perspective for training PINNs on strongly coupled spatial PDE problems.

\section{Related Works}
\label{sec:rel}

\subsection{Information Propagation Hypothesis}

Beyond non-convex loss landscapes and loss-term imbalance, a growing line of work interprets PINN training through the \emph{information propagation hypothesis}: correct solutions must effectively transmit constraint information from initial/boundary conditions into the interior to form a globally consistent field, yet this transmission can break down even when the overall loss decreases. 
Penwarden et al. systematically analyze this phenomenon and categorize propagation failure into three representative modes: \emph{zero solution}, where large interior regions collapse to near-trivial states and boundary information is not injected; \emph{no propagation}, where inadequate constraints or poor sampling prevent boundary information from reaching the interior; and \emph{incorrect propagation}, where local structures, strong coupling, or multi-scale stiffness block transmission and induce wrong solution forms in subregions, ultimately destroying global consistency.\textsuperscript{\cite{penwarden2023unified}} 
Building on this view, Daw et al.\ show that propagation failure may manifest as interior residual barriers and can be diagnosed via highly imbalanced residual distributions (e.g., skewness and kurtosis), and they propose a 3R (retain--resample--release) strategy to mitigate it.\textsuperscript{\cite{daw2022mitigating}} 
Relatedly, Wang et al.\ frame time-domain extrapolation as sequential information propagation and introduce a gated parameter-space correction term to extend a pretrained solution from a short interval to a larger one.\textsuperscript{\cite{wang2025extrapolation}}

\subsection{Curriculum Learning}

Curriculum learning trains PINNs from easy to hard, letting the network fit simpler structures first and then progressively handle more complex ones, which stabilizes early optimization and reduces poor local minima.
Some studies analyze failure mechanisms and show that PINNs are vulnerable to blocked information propagation and multi-objective conflicts, motivating curriculum or progressive training to improve convergence stability\textsuperscript{\cite{krishnapriyan2021characterizing}}. In terms of concrete designs, curricula have been constructed from a spatial-subdomain perspective by partitioning the domain into subregions, grouping PDE losses, and adaptively reweighting them based on residual magnitudes to ease simultaneous global optimization\textsuperscript{\cite{cao2025adaptive}}. Other works adopt a constraint-first and refinement-later strategy: the network is initialized with partial boundary conditions or easier PDE constraints, and then stricter constraints are gradually introduced to refine the solution\textsuperscript{\cite{jahani2024enhancing}}. Curricula can also be designed via data or sampling resolution (e.g., switching among multi-density point sets to learn low- and high-frequency components)\textsuperscript{\cite{tsai2025mld}}, or along the parameter dimension by progressively increasing parameter complexity with adaptive step control\textsuperscript{\cite{duffy2024dynamic}}. Finally, dynamic sampling strategies gradually expand the sampling distribution to improve global coverage and promote information propagation\textsuperscript{\cite{munzer2022curriculum}}.
A systematic comparison and analysis of curriculum learning, causality-guided training, and Fourier-feature-based methods has been provided in the literature\textsuperscript{\cite{monaco2023training}}.

\subsection{Causality-guided Training}

For transient PDEs with explicit time-causality structures, causality-guided methods use "early-to-late" stepwise training to transform the global problem into a series of subproblems on time slices/windows, thus avoiding erroneous propagation due to causality violations. 
PT-PINN performs time-local pretraining on a short window and then extends to the full horizon by parameter continuation with pseudo-supervision\textsuperscript{\cite{guo2023pre}}. TsoNN mitigates ill-conditioning via time stepping over successive time intervals\textsuperscript{\cite{cao2023tsonn}}.
Mattey \emph{et al.} progressively enlarge the time window, generate pseudo-labels on previously trained regions, and use transfer learning to adapt the model to newly added intervals\textsuperscript{\cite{mattey2022novel}}. 
Penwarden \emph{et al.} provide a unified view of causality-guided paradigms and propose an integrated framework to improve stability and applicability\textsuperscript{\cite{penwarden2023unified}}.
Related strategies further organize training by categorizing subregions as trained/being-trained/to-be-trained based on PDE loss and weighting profiles\textsuperscript{\cite{lin2025causality}}. 
For long-time integration, dual-network or implicit-propagation ideas and sliding-window continuation have been explored to enhance stability\textsuperscript{\cite{guo2025dual,guo2025long}}. 
In stiff or singularly perturbed settings, improving the accuracy of initial constraints and applying time-dependent reweighting (e.g., negative-exponential weights driven by accumulated error) can help suppress error growth\textsuperscript{\cite{cao2024multistep,WANG2024116813}}.

\section{Method}

\begin{figure}[!htbp]
    \centering
    \includegraphics[width=\linewidth]{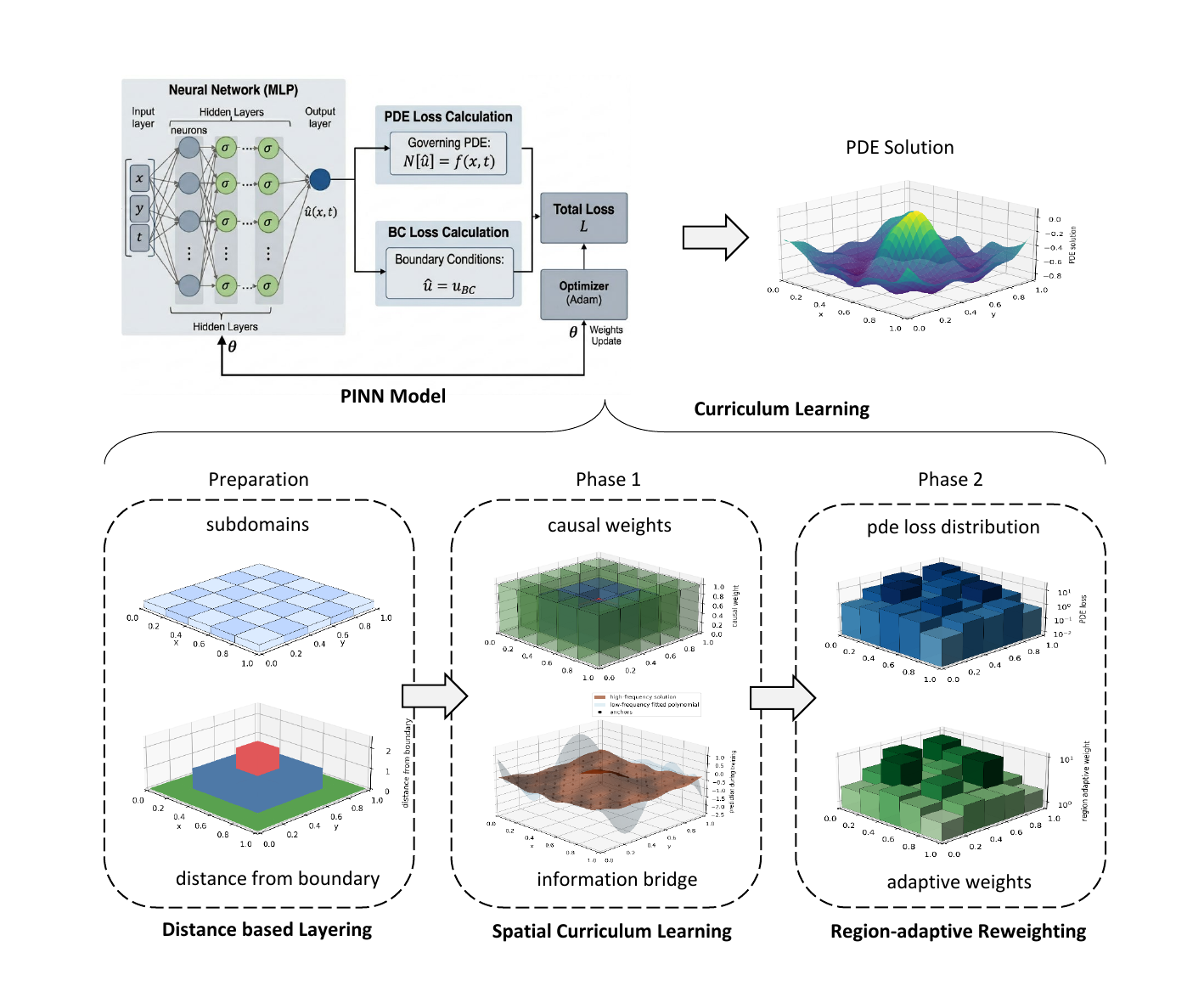}
    \caption{\textbf{Overview of the proposed method.} 
    The upper-left panel shows the classical PINN model. 
    The proposed curriculum learning framework consists of three main parts. 
    As a preparation step, the computational domain is partitioned into multiple subregions and further organized into distance-based layers from the boundary toward the interior. 
    In the first phase, a spatially correlated curriculum is implemented by assigning causal weights to the PDE losses of different layers. Meanwhile, an information bridge based on a low-order fitting function is introduced to capture the global low-frequency component.
    In the second phase, a region-adaptive reweighting strategy is introduced to dynamically rebalance the PDE loss weights across subregions, so that insufficiently optimized regions receive appropriate training emphasis and local accuracy is further improved.
    Through these stages, the network progressively recovers the solution from a globally consistent low-frequency structure to accurate local high-frequency details.}
    \label{fig:overview}
\end{figure}

In this section, we present the overall framework of the proposed method. 
As illustrated in Fig.~\ref{fig:overview}, our approach extends the classical PINN paradigm by introducing a spatially correlated curriculum strategy for training over the computational domain. 
The whole framework is organized into a preparation step and two successive stages: distance-based domain layering, spatially correlated curriculum learning with an information bridge, and region-adaptive local reweighting.
These components work together to guide the network from learning a globally consistent low-frequency structure to progressively recovering accurate local high-frequency details.

\subsection{Spatial curriculum learning}
For boundary value problems (BVPs), the solution is determined by appropriate and sufficient boundary conditions rather than by a strict temporal causality structure. For more details, one may refer to Appendix A, which discusses the mathematical classification of PDEs, their analysis from the perspective of physical propagation, and the characteristics of BVPs.

In the absence of additional supervised data, PINNs can only determine the global solution by combining boundary conditions and PDE constraints. 
However, in many strongly coupled or nonlinear, multi-scale PDEs, boundary information is not automatically and consistently propagated into the domain during training: 
on one hand, a single boundary is often insufficient to uniquely determine the solution near it; 
On the other hand, the global solution structure is governed by the coupling between boundary conditions and PDE operators.
As discussed in the related work, such coupling may make simultaneous optimization over the entire domain prone to poor local minima and ineffective information propagation.

To address this, we propose a spatial weak-causality constraint curriculum training strategy: by using soft constraints to reinforce the PDE residual constraints near the boundaries, we guide the network's training focus to gradually move from the boundaries inward, thereby reducing the overall optimization difficulty.

\begin{figure}[htbp]
\centering
  \subfigure[1D domain]{\includegraphics[width=0.6\linewidth]{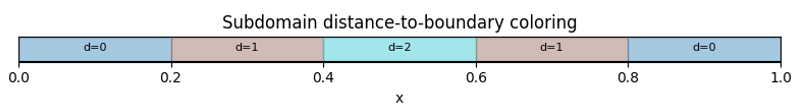}}
  \subfigure[2D domain]{\includegraphics[width=0.4\linewidth]{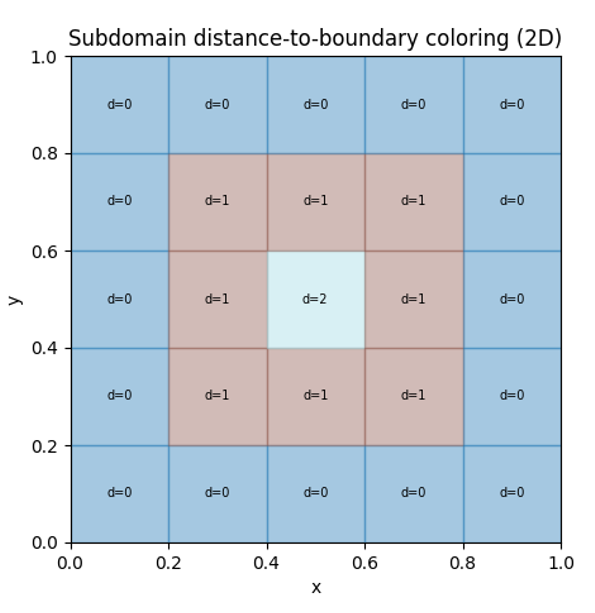}}
  \caption{ \textbf{Boundary distance based layering in 1D and 2D domains.} Here, $d$ denotes the discrete distance to the boundary, and both examples can be partitioned into three layers with $d=0,1,2$.}
  \label{fig:layer_distance}
\end{figure}

\paragraph{Layered Progression from Boundaries Inward.}
We divide the solution domain $\Omega$ into multiple discrete subregions and organize them into several layers according to their distances to the boundary, as shown in Fig.~\ref{fig:layer_distance}, which illustrates the layered structures for both 1D and 2D solution domains. Intuitively, the network should first capture the solution near the boundary, and then the training focus gradually extends to deeper regions.

Unlike time-causality-guided methods, this approach does not require strict unidirectional causality propagation in space. Instead, we impose a weak spatial causality bias: learning begins at the outer layers and progressively moves inward. If the outer layers are not learned correctly, the inner layers are also likely to be inaccurate.

Let $\partial \Omega$ denote the boundary of the domain. We first partition $\Omega$ into $N$ discrete subregions, denoted by $\{\Omega_i\}_{i=1}^N$. For each subregion $\Omega_i$, we define its distance to the boundary as
\begin{equation}
    d_i = \mathrm{dist}(\Omega_i,\partial\Omega),
\end{equation}
where the distance can be evaluated, for example, by the minimum distance between $\Omega_i$ and $\partial\Omega$. Based on these distances, the subregions are grouped into $L$ layers from the outer region to the inner region.

In the 1D case, the interval is partitioned into several discrete subregions and grouped into three layers from the boundary to the center, while in the 2D case, the square domain is partitioned into multiple discrete subregions and organized into concentric layers according to their distances to the boundary.

\paragraph{Cumulative negative exponential weighting}
Inspired by existing causality-guided PINN methods for time-dependent problems, directional solution fitting can be encouraged by introducing negatively exponentially decayed weights\textsuperscript{\cite{WANG2024116813}}. 

Specifically, the time domain is divided into multiple intervals, and the cumulative PDE loss is evaluated sequentially to adaptively adjust the causal weights distribution. As a result, later intervals are emphasized only after earlier ones have been sufficiently optimized, which helps prevent premature convergence in later intervals from interfering with the training of earlier ones.

Based on this, we divide the domain $\Omega$ into $L$ layers from the outer region to the inner region, and let $\Lambda_j$ denote the set of sampling points in the $j$-th layer. The PDE loss associated with the $j$-th layer is defined as
\begin{equation}
  \mathcal{L}^{(j)}_{\mathrm{PDE}}
  = \frac{1}{N_j}\sum_{\mathbf{x}\in \Lambda_j}
  \left| \mathcal{N}(u_\theta(\mathbf{x})) \right|^2,
\end{equation}
where $N_j = |\Lambda_j|$ is the number of sampled points in the $j$-th layer, $u_\theta$ is the PINN approximation, and $\mathcal{N}(\cdot)$ denotes the PDE differential operator, whose derivatives are computed via automatic differentiation.

\begin{figure}[htbp]
\centering
  \subfigure[1D domain]{\includegraphics[width=0.35\linewidth]{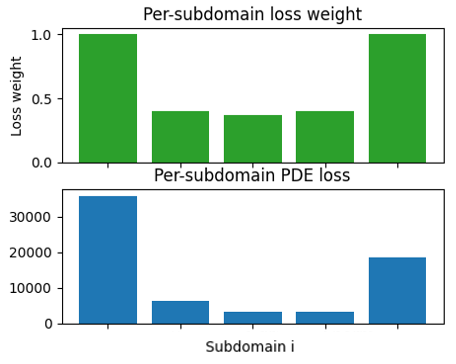}}
  \subfigure[2D domain]{\includegraphics[width=0.6\linewidth]{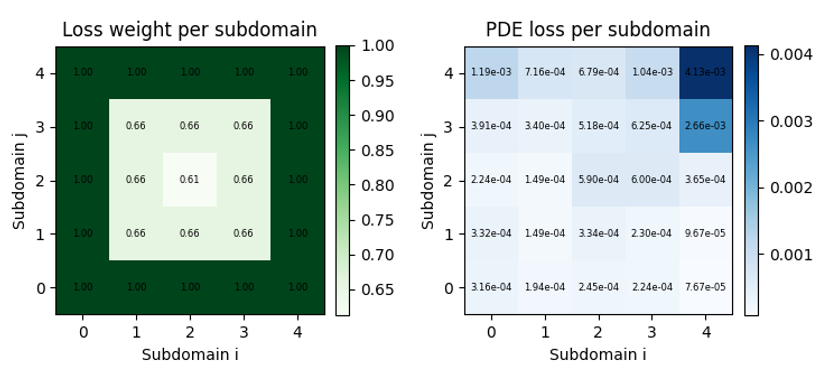}}
  \caption{\textbf{Causal-weight and subregional PDE-loss distributions in the 1D and 2D cases.} The causal weights decrease from the boundary toward the interior and are adaptively adjusted according to the PDE-loss distribution across subregions.}
  \label{fig:causal_weight}
\end{figure}

To reflect the boundary-to-interior curriculum progression, we assign a dynamic weight $w_i$ to the PDE loss of the $i$-th layer, such that the outer layers are emphasized at the early stage of training. A simple and effective choice is an exponentially decayed weight based on the accumulated residuals of the preceding outer layers:
\begin{equation}
  w_i = \exp\left(-\epsilon \sum_{j=0}^{i-1} \mathcal{L}^{(j)}_{\mathrm{PDE}} \right),
  \label{eq:layer_weight}
\end{equation}
where $\epsilon > 0$ is the decay parameter. This design has a clear gating effect: when the residuals in the outer layers remain large, $\sum_{j=0}^{i-1} \mathcal{L}^{(j)}_{\mathrm{PDE}}$ is also large, which significantly suppresses the weights of the $i$-th layer. Only after the outer layers gradually converge do the weights of the inner layers increase, allowing the training focus to progressively shift toward the interior regions. 
In this way, the proposed mechanism numerically realizes a weak causality process in which the outer layers are fitted first and the inner layers are learned progressively afterward. Figure~\ref{fig:causal_weight} illustrates the distributions of the causal weights and PDE losses for the 1D and 2D cases.

The final weighted PDE loss is written as
\begin{equation}
  \mathcal{L}_{\mathrm{PDE}} = \sum_{i=0}^{L-1} w_i \mathcal{L}^{(i)}_{\mathrm{PDE}},
\end{equation}
and it is combined with the boundary condition loss $\mathcal{L}_{\mathrm{BC}}$ (and possibly data loss $\mathcal{L}_{\mathrm{data}}$) to form the total objective:
\begin{equation}
  \mathcal{L} = \lambda_{\mathrm{BC}} \mathcal{L}_{\mathrm{BC}} + \lambda_{\mathrm{data}} \mathcal{L}_{\mathrm{data}} + \mathcal{L}_{\mathrm{PDE}}.
\end{equation}

\subsection{Information Bridge}

In boundary value problems (BVPs), the global solution structure is usually shaped by the joint effect of multiple boundary conditions rather than by any single boundary alone. 
As a result, accurately fitting local boundary regions does not necessarily ensure global consistency. 
The solution patterns induced by different boundaries must be coordinated within the interior domain through the governing PDE.

However, during PINN training, inaccurate approximations in the intermediate regions may weaken the effective coupling of boundary-induced solution structures, particularly in regions with high-frequency oscillations or persistent local residuals. 
This can create an information barrier that prevents boundary constraints from being consistently reconciled across the domain. 
Consequently, the learned solution may suffer from inconsistent cross-region fitting and pronounced low-frequency drift, including shifts in the overall trend, baseline errors, and phase or amplitude deviations, even when local residuals are reduced.

\begin{figure}[htbp]
\centering
  \subfigure[Low frequency drift]{\includegraphics[width=0.6\linewidth]{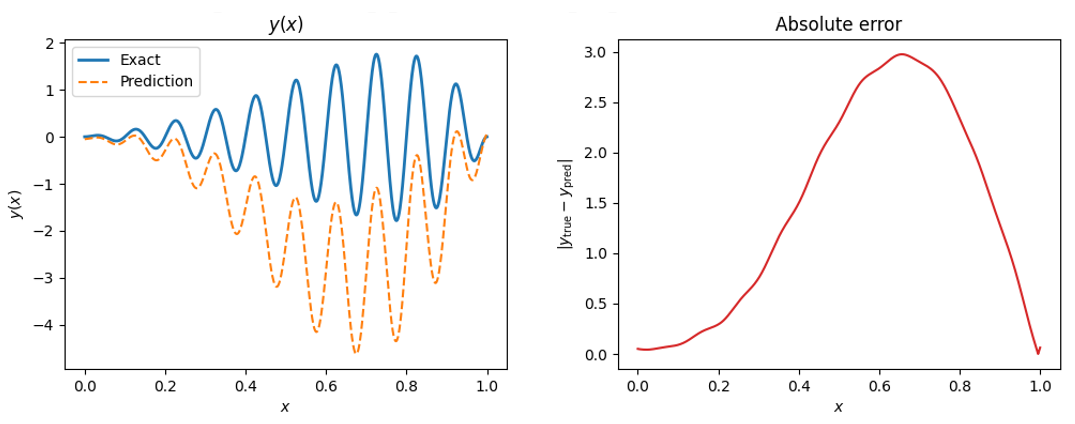}}
  \subfigure[Low order fitting]{\includegraphics[width=0.34\linewidth]{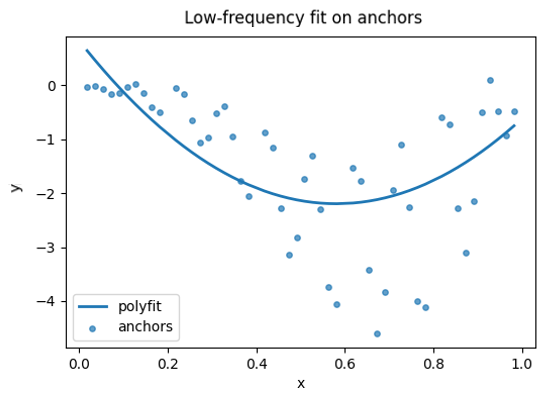}}
  \caption{\textbf{Low frequency drift and weighted low order fitting.} (a) shows the 1D ODE example used in the later experiments. In this case, the training process is dominated by the PDE loss, so the high-frequency structure of the solution is captured reasonably well, while a significant low-frequency drift still remains, as the solutions near the two boundaries fail to converge consistently to a globally compatible solution. 
  (b) illustrates the weighted low-order fitting performed on a set of selected anchor points, yielding a function that represents the low-frequency component of the solution.}
  \label{fig:low_fre_fit}
\end{figure}

To mitigate this issue, we design a low-frequency information bridge mechanism: by using a few sparse anchor points to extract reliable low-frequency components from the current network prediction, we impose a consistency constraint between distant boundaries through a "soft" connection, thus stabilizing the global structure early in training and reducing the risk of low-frequency drift, as shown in Fig.~\ref{fig:low_fre_fit}(a).

\paragraph{Low-order fitting with anchors.}

Let the current PINN prediction be $u_\theta(\mathbf{x})$. We select a set of anchor points $\mathcal{A}=\{\mathbf{x}_i\}_{i=1}^N$, which are typically placed in the intermediate region or other key locations that connect different subregions, and perform a weighted fitting of the predicted values at these anchor points using low-order functions to extract the low-frequency component. For example, in the 1D case with a second-order polynomial basis, we define
\begin{equation}
  \phi(x) = [1,\ x,\ x^2]^{\top}, \qquad \hat u_{\mathrm{LF}}(x;\beta) = \phi(x)^{\top}\beta .
\end{equation}
The corresponding weighted least-squares estimate is
\begin{equation}
  \beta^{\ast} = \arg\min_{\beta} \sum_{i=1}^{N} \alpha_i \left( \phi(x_i)^{\top}\beta - \hat u_i \right)^2,
  \label{eq:wls_polyfit}
\end{equation}
where $\hat u_i = u_\theta(x_i)$ denotes the network prediction at the anchor point $x_i$, and $\alpha_i$ is the weight associated with that anchor point. For the 2D case, higher-order polynomials or other low-frequency bases, such as cubic polynomials or radial basis functions, can be adopted in the same manner by extending $\phi(\cdot)$ to two-dimensional basis functions.

This procedure extracts a globally smooth trend from the current network prediction and uses it as a shared low-frequency prior across different regions, as illustrated in Fig.~\ref{fig:low_fre_fit}(b). Intuitively, when high-frequency errors remain significant in the intermediate regions, directly enforcing cross-region consistency through the PDE residual can be ineffective. In contrast, low-frequency trends are more stable and easier to learn, and thus provide a suitable consistency bridge between boundary-induced solution structures.

\paragraph{Information bridge regularization using pseudo labels.}

After obtaining the low-frequency fitting function $\hat u_{\mathrm{LF}}$, we introduce a low-frequency consistency loss over the bridge anchors as a regularization term:
\begin{equation}
  \mathcal{L}_{\mathrm{LF}} = \mathbb{E}_{\mathbf{x} \sim \mathcal{A}} \left[ u_\theta(\mathbf{x}) - \hat u_{\mathrm{LF}}(\mathbf{x}) \right]^2,
  \label{eq:lf_loss}
\end{equation}
where $\hat u_{\mathrm{LF}}(\mathbf{x})$ serves as a pseudo-label generated by the low-frequency fitting, and $\mathcal{A}$ denotes the anchor set introduced above. The total loss is then written as
\begin{equation}
  \mathcal{L}_{\mathrm{total}} = \mathcal{L}_{\mathrm{PDE}} + \mathcal{L}_{\mathrm{BC}} + \lambda_{\mathrm{LF}} \mathcal{L}_{\mathrm{LF}},
  \label{eq:total_with_lf}
\end{equation}
where $\lambda_{\mathrm{LF}}$ controls the strength of the information bridge. 
The loss term in Eq.~\eqref{eq:lf_loss} acts as a soft regularization constraint. 
It does not replace the PDE and boundary constraints, but instead provides stable low-frequency guidance during training, encouraging the network to establish a globally consistent baseline and a coherent slowly varying structure across different subregions.

The information bridge is mainly applied to regions that have not yet been sufficiently optimized by the curriculum training. 
Since the layered PDE weights are smaller in these regions, the corresponding pseudo-label weights are set to be larger. 
This complementary weighting scheme allows the low-frequency prior to provide additional guidance where the PDE residual constraint is still relatively weak. 
As training proceeds, the PDE weights of these regions gradually increase, and the role of the bridge is correspondingly reduced, so that the final solution is still primarily governed by the PDE and boundary constraints rather than by the pseudo-labels.

The reason for referring to this mechanism as an information bridge is that the extracted low-frequency trend provides a shared reference across spatially separated regions. 
When the PDE residual alone cannot yet establish consistent coupling between distant boundaries, this shared low-frequency prior helps connect boundary-induced solution structures through the interior domain. 
By constraining the global baseline and slowly varying components of the solution, the bridge reduces low-frequency drift and improves cross-region consistency. 
It complements the spatial curriculum training in Section~3.1: the curriculum controls the boundary-to-interior training order, whereas the information bridge maintains low-frequency coherence across regions during early training.

\subsection{Region-adaptive Reweighting}

After applying the spatial curriculum learning and the information bridge constraint introduced above, the PINN is generally less prone to training failure and can obtain a reasonably accurate solution.

However, the optimization states of different subregions may still be highly unbalanced. 
Although the first phase can produce a reasonably good global approximation, some regions may remain insufficiently optimized, especially those with strong high-frequency variations, complex boundary influences, or insufficient information propagation. 
These regions may still exhibit large local PDE residuals and noticeable local errors.

\paragraph{Gradient-aware region-adaptive residual reweighting.}

The difficulty of optimizing these regions is not determined solely by
the magnitude of their residuals.
In some subregions, the PDE loss may remain relatively large, while the
corresponding loss term produces only weak gradients with respect to the network
parameters during training.
Consequently, these difficult regions may receive insufficient effective updates
despite their large residuals.

To alleviate this issue, we introduce a gradient-aware region-adaptive reweighting mechanism that recalibrates the PDE residual weights by jointly considering the magnitude of the regional PDE loss and its corresponding gradient norm.
Similar local reweighting strategies adjust regional weights according to local PDE residuals\textsuperscript{\cite{cao2025adaptive}}, but they do not explicitly account for the gradient norm.

Suppose that the solution domain is partitioned into \(N\) subregions
\(\{\Omega_i\}_{i=1}^N\), and let
\(\mathcal{L}_{\mathrm{PDE}}^{(i)}\) denote the PDE loss evaluated on the
\(i\)-th subregion.
For each subregion, we compute the gradient norm of its PDE loss with respect
to the network parameters:
\begin{equation}
G_i
=
\left\|
\nabla_{\theta}\mathcal{L}_{\mathrm{PDE}}^{(i)}
\right\|_2
=
\left(
\sum_{\ell}
\left\|
\frac{\partial \mathcal{L}_{\mathrm{PDE}}^{(i)}}{\partial \theta_{\ell}}
\right\|_2^2
\right)^{1/2},
\qquad i=1,\dots,N.
\label{eq:regional_grad_norm}
\end{equation}
where \(\theta=\{\theta_{\ell}\}_{\ell}\) denotes all trainable parameters of
the neural network.
The quantity \(G_i\) measures how strongly the PDE loss in the \(i\)-th
subregion contributes to the parameter updates.
Motivated by the neural tangent kernel (NTK)
perspective\textsuperscript{\cite{wang2022and}}, we use this gradient norm as
a practical indicator to quantify the influence of each regional residual on
the training dynamics.
A smaller \(G_i\) suggests that the corresponding regional residual has a weaker
effect on parameter updates, even when its loss magnitude remains large.

We then define a regional difficulty score by combining the PDE loss magnitude
and the gradient norm:
\begin{equation}
s_i
=
\log\left(\mathcal{L}_{\mathrm{PDE}}^{(i)}+\varepsilon\right)
-
\lambda_g
\log\left(G_i+\varepsilon\right),
\qquad i=1,\dots,N,
\label{eq:regional_difficulty_score}
\end{equation}
where \(\varepsilon>0\) is a small constant for numerical stability, and
\(\lambda_g\geq 0\) controls the relative contribution of the gradient-norm
term.
When \(\lambda_g=0\), the score is determined only by the regional PDE loss,
and the reweighting mechanism mainly emphasizes high-residual regions.
As \(\lambda_g\) increases, the score assigns greater importance to regions
with weaker gradient contributions.
Therefore, regions with large PDE losses and small gradient norms receive larger
difficulty scores, indicating that they are both poorly fitted and insufficiently
optimized.

The regional coefficients are obtained by mapping the difficulty scores into a
bounded interval:
\begin{equation}
\rho_i
=
\Phi(s_i),
\qquad
\Phi:\mathbb{R}\rightarrow [\rho_{\min},\rho_{\max}],
\label{eq:regional_weight_mapping}
\end{equation}
where \(0<\rho_{\min}<\rho_{\max}\) are prescribed constants.
The mapping \(\Phi(\cdot)\) is chosen to be monotone nondecreasing, so that
larger difficulty scores lead to larger regional weights.
In practice, \(\Phi(\cdot)\) can be implemented by min--max normalization
followed by linear rescaling:
\begin{equation}
\tilde{s}_i
=
\frac{s_i-\min_j s_j}
{\max_j s_j-\min_j s_j+\varepsilon},
\qquad
\rho_i
=
\rho_{\min}
+
(\rho_{\max}-\rho_{\min})\tilde{s}_i.
\label{eq:minmax_regional_weight}
\end{equation}
This bounded mapping prevents excessive disparities among regional weights and
improves the stability of the adaptive reweighting process.
For example, the interval \([\rho_{\min},\rho_{\max}]\) can be set to \([1,5]\).

\begin{figure}[htbp]
\centering
  \includegraphics[width=0.9\linewidth]{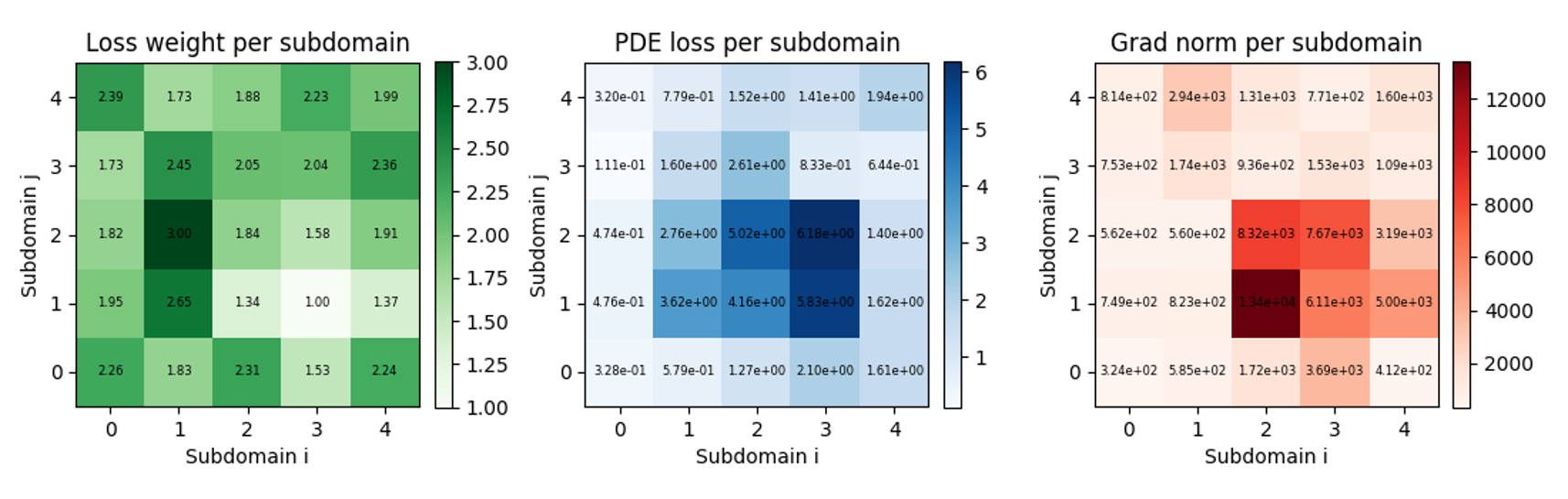}
  \caption{\textbf{Illustration of the region-adaptive reweighting mechanism.}
  Adaptive regional weights, PDE losses, and gradient magnitudes across subregions.
  The proposed strategy assigns larger weights to regions with large residuals but weak gradient effectiveness.}
  \label{fig:adaptive_weight}
\end{figure}

Based on these regional coefficients, the adaptively weighted PDE loss is
defined as
\begin{equation}
\mathcal{L}_{\mathrm{PDE}}^{\mathrm{adapt}}
=
\frac{1}{N}
\sum_{i=1}^{N}
\rho_i\,
\mathcal{L}_{\mathrm{PDE}}^{(i)}.
\label{eq:adaptive_pde_loss}
\end{equation}
In this way, the proposed strategy does not merely emphasize regions with large
residuals.
Instead, it dynamically rebalances the training process by assigning greater
weights to regions that are both difficult to fit and insufficiently influential
in the parameter-update dynamics.
A representative example is shown in Fig.~\ref{fig:adaptive_weight}, where the PDE loss, gradient norm, and corresponding loss weight of each subregion are visualized.

\section{Experiment}

The experiments were conducted in a Linux environment equipped with an NVIDIA RTX 5090 GPU (32 GB) and an Intel Xeon Gold 6348 processor with 128 GB of memory, using Python libraries including PyTorch, NumPy, and DeepXDE.

In this study, we compared the proposed PINN with spatially correlated curriculum training (PINN-C) against the classical PINN model. 
We consider one ODE example as an introductory pedagogical case, together with three representative PDE problems.
All models were trained using the same network architecture, the same number of sampling points, and the same optimizer settings (the detailed hyperparameters are provided in Appendix C). 
In the first phase, causal weights and a low-frequency information bridge were introduced to facilitate information propagation across the domain. 
In the second phase, region-adaptive reweighting was further incorporated to improve the accuracy of the final solution. 
Here, PINN-C1 denotes the Phase-I-only variants, with PINN-C1(with bridge) and PINN-C1(w/o bridge) indicating whether the information bridge is used. PINN-C2 denotes the full method using both Phase I and Phase II.

Each experiment is repeated three times with different random seeds.
The models are optimized sequentially using Adam and LBFGS.
Intermediate results are recorded at fixed intervals, based on which the training weights are adjusted and the low-frequency fitting function is updated.
The collocation points are uniformly distributed over the domain.
To evaluate the model performance, we analyzed both the training loss curves and the test error curves, using metrics including the mean squared error (MSE), $L_2$ relative error, maximum absolute error, and average PDE residual. 
In addition, we visualized the absolute error and its spatial distribution over the domain. 
The test data used for validation were either generated from analytical solutions or obtained from COMSOL solvers. 
Four types of PDE problems were considered in the experiments, and their detailed mathematical formulations are provided in Appendix B. 
The code has been publicly released at \url{https://github.com/pigofmomo/CurriculumLearningPINN}.

\subsection{1D ODE}

This problem serves as an introductory pedagogical example, from which the proposed spatial curriculum learning strategy was first motivated, analyzed, and validated.
It considers a one-dimensional ODE with Dirichlet boundary conditions imposed at both endpoints.
The differential operator contains both an exponentially growing term and an oscillatory term.
Through a careful construction, the problem admits an analytical solution.
We consider two variants, corresponding to low-frequency and high-frequency settings.

For the curriculum design, the domain is partitioned into five subregions, which are further organized into three layers from the boundary toward the interior.
Due to the limitations of network capacity and sampling density, we do not further apply region-adaptive reweighting to improve local accuracy in this example.
Under the current configuration, the obtained accuracy is already satisfactory and is close to the performance limit of the model setup.
The quantitative results over multiple runs are reported in Table~\ref{tab:ode1d_results}.
 
\begin{table}[htbp]
\centering
\caption{Quantitative results for 1D ODE problems under low- and high-frequency settings.}
\label{tab:ode1d_results}
\resizebox{\linewidth}{!}{%
\begin{tabular}{|l|l|c|c|c|c|}
\hline
\text{Frequency} & \text{Method} & \text{PDE Residual} & $L_2$ Relative Error & \text{Max Abs Error} & \text{MSE} \\
\hline
\multirow{3}{*}{Low}
& PINN (default BC weight)
& $(9.22 \pm 5.83)\times 10^{0}$
& $(1.52 \pm 0.09)\times 10^{0}$
& $(1.06 \pm 0.06)\times 10^{0}$
& $(3.76 \pm 0.44)\times 10^{-1}$ \\
\cline{2-6}
& PINN (strong BC weight)
& $(2.27 \pm 1.57)\times 10^{2}$
& $(6.85 \pm 4.45)\times 10^{-1}$
& $(5.77 \pm 3.74)\times 10^{-1}$
& $(1.08 \pm 0.76)\times 10^{-1}$ \\
\cline{2-6}
& PINN-C1 (w/o bridge)
& $\mathbf{(7.05 \pm 3.22)\times 10^{0}}$
& $\mathbf{(5.70 \pm 0.58)\times 10^{-2}}$
& $\mathbf{(5.16 \pm 0.62)\times 10^{-2}}$
& $\mathbf{(5.34 \pm 0.01)\times 10^{-4}}$ \\
\hline
\multirow{3}{*}{High}
& PINN (strong BC weight)
& $(1.60 \pm 1.13)\times 10^{3}$
& $(6.78 \pm 4.56)\times 10^{-1}$
& $(1.21 \pm 0.82)\times 10^{0}$
& $(4.08 \pm 2.88)\times 10^{-1}$ \\
\cline{2-6}
& PINN-C1 (w/o bridge)
& $(6.45 \pm 6.70)\times 10^{1}$
& $(1.47 \pm 1.01)\times 10^{0}$
& $(1.82 \pm 1.31)\times 10^{0}$
& $(1.94 \pm 2.06)\times 10^{0}$ \\
\cline{2-6}
& PINN-C1 (with bridge)
& $\mathbf{(1.75 \pm 0.45)\times 10^{0}}$
& $\mathbf{(3.50 \pm 0.08)\times 10^{-2}}$
& $\mathbf{(7.07 \pm 0.14)\times 10^{-2}}$
& $\mathbf{(7.49 \pm 0.03)\times 10^{-4}}$ \\
\hline
\end{tabular}%
}
\end{table}

\begin{figure}[!htbp]
\centering
  \subfigure[]{\includegraphics[width=0.4\linewidth]{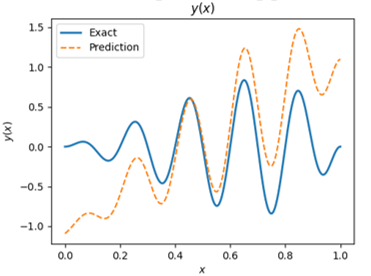}}\hspace{0.05\linewidth}
  \subfigure[]{\includegraphics[width=0.3\linewidth]{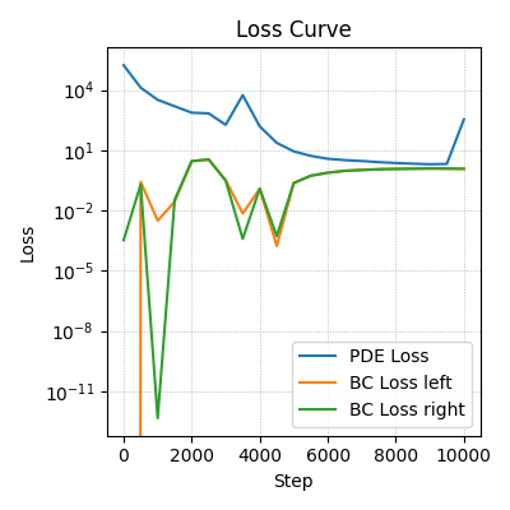}}
  \subfigure[]{\includegraphics[width=0.4\linewidth]{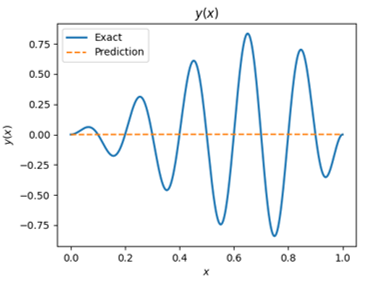}}\hspace{0.05\linewidth}
  \subfigure[]{\includegraphics[width=0.3\linewidth]{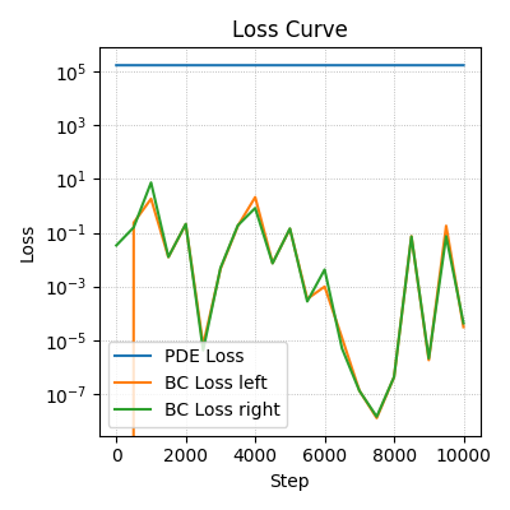}}
  \subfigure[]{\includegraphics[width=0.4\linewidth]{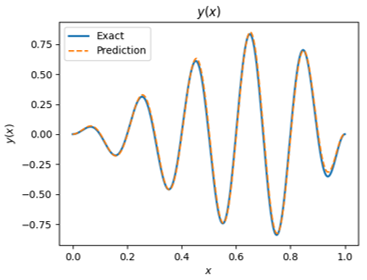}}\hspace{0.05\linewidth}
  \subfigure[]{\includegraphics[width=0.3\linewidth]{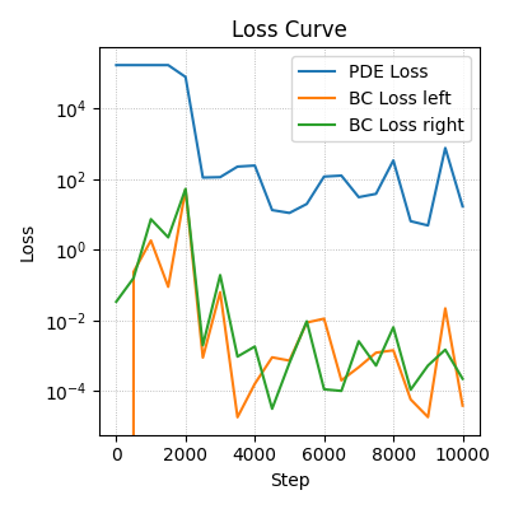}}
  \caption{\textbf{Results for the low-frequency 1D ODE case.}
          (a)(b) Classical PINN with the default loss weights, where the PDE loss dominates and the boundary conditions are not satisfied.
          (c)(d) PINN with a strong BC weight, which collapses to a trivial zero solution due to failed information propagation.
          (e)(f) PINN-C1 with causal weights only, which successfully guides boundary information into the interior and recovers the correct solution.
          }
  \label{fig:ode_low}
\end{figure}

\paragraph{Low frequency case.}
We first consider the classical PINN with the default PDE loss weight set to 1, and the result is shown in Fig.~\ref{fig:ode_low}(a).
Although the high-frequency structure in the middle region is partially fitted, the overall solution is incorrect because the boundary conditions at both endpoints are not satisfied.
This failure can be attributed to the dominance of the PDE loss during training, while the BC constraint receives insufficient emphasis because of its relatively small weight.
As shown by the loss curves in Fig.~\ref{fig:ode_low}(b), although the PDE loss decreases throughout training, the BC loss fails to converge.
This observation further highlights that satisfying the boundary conditions is essential for recovering the correct solution.

We then consider a PINN with a strong boundary constraint, where a large BC weight of 100 is applied, and the result is shown in Fig.~\ref{fig:ode_low}(c).
However, the model instead converges to a trivial zero solution, indicating a failure of information propagation.
Although the boundary conditions are enforced, the PDE loss remains difficult to optimize because of its stiffness, preventing the boundary information from propagating into the interior, as reflected in Fig.~\ref{fig:ode_low}(d).

Next, we introduce the proposed spatial curriculum learning strategy using only the causal weights, without the information bridge at this stage.
The result, shown in Fig.~\ref{fig:ode_low}(e), successfully recovers the correct solution.
The loss curves in Fig.~\ref{fig:ode_low}(f) further show that all terms can gradually converge during training.
This demonstrates that the introduction of causal weights can effectively guide information propagation, reduce the optimization difficulty, and alleviate the conflict between the boundary loss and the PDE loss. 
As shown in Table~\ref{tab:ode1d_results}, the quantitative accuracy of PINN-C1 (w/o bridge) is substantially better than that of the two preceding methods.

\paragraph{High frequency case.}
We increase the exponential growth rate and the sinusoidal oscillation frequency to construct a high-frequency case for further investigation.
In this setting, the oscillatory behavior becomes more pronounced and the PDE operator exhibits stronger stiffness.

\begin{figure}[!htbp]
\centering
  \subfigure[]{\includegraphics[width=0.4\linewidth]{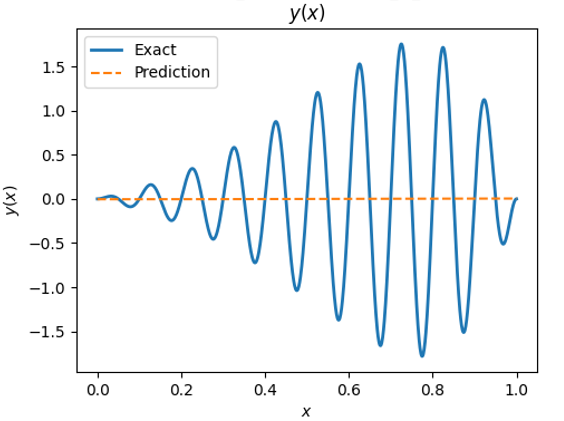}}\hspace{0.05\linewidth}
  \subfigure[]{\includegraphics[width=0.3\linewidth]{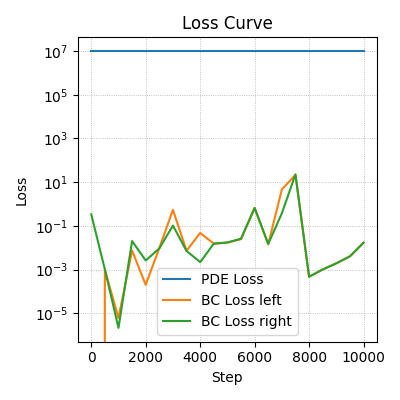}}
  \subfigure[]{\includegraphics[width=0.4\linewidth]{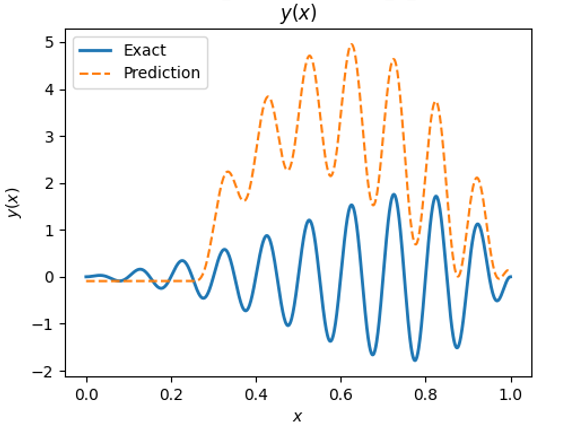}}\hspace{0.05\linewidth}
  \subfigure[]{\includegraphics[width=0.3\linewidth]{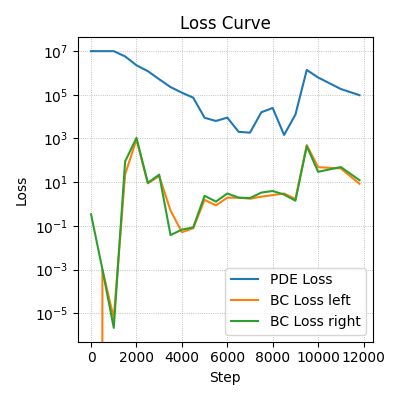}}
  \subfigure[]{\includegraphics[width=0.4\linewidth]{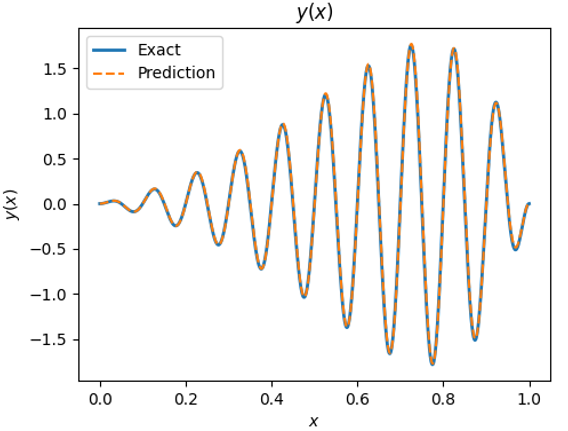}}\hspace{0.05\linewidth}
  \subfigure[]{\includegraphics[width=0.3\linewidth]{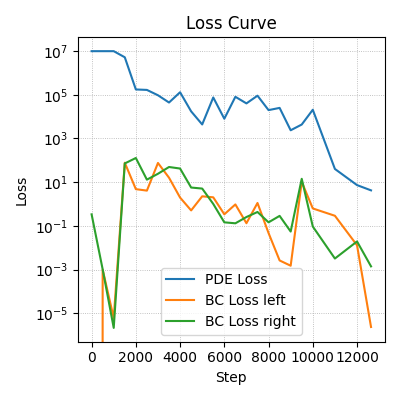}}
  \caption{\textbf{Results for the high-frequency 1D ODE case.}
          (a)(b) PINN with a strong BC weight, which reduces the BC loss but collapses to a trivial zero solution due to failed information propagation.
          (c)(d) PINN-C1 without the information bridge, which recovers the main solution structure but still exhibits low-frequency drift.
          (e)(f) PINN-C1 with the information bridge, which suppresses low-frequency drift and improves global consistency, leading to a more accurate solution.}
  \label{fig:ode_high}
\end{figure}

We first consider the PINN with a strong BC weight of 1000.
As shown in Fig.~\ref{fig:ode_high}(a), the model again converges to a trivial zero solution, indicating a failure of information propagation.
Although the BC loss is reduced to a very small value, the PDE loss remains difficult to optimize (see Fig.~\ref{fig:ode_high}(b)).

\begin{figure}[!htbp]
\centering
  \subfigure[]{\includegraphics[width=0.4\linewidth]{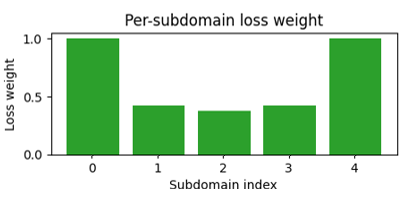}}\hspace{0.05\linewidth}
  \subfigure[]{\includegraphics[width=0.3\linewidth]{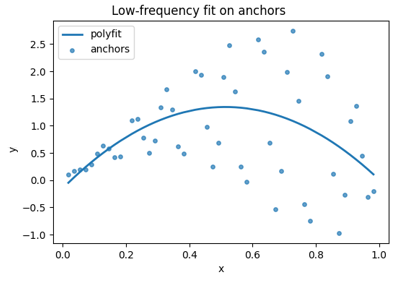}}
  \caption{\textbf{Illustration of the training process for the high-frequency 1D ODE case.}
  (a) Boundary-to-interior causal weighting.
  (b) Sampled anchor points and the fitted second-order polynomial used in the information bridge.}
  \label{fig:ode_high_causal}
\end{figure}

Next, we incorporate causal weights into the proposed spatial curriculum learning framework, assigning larger weights to regions near the boundaries and smaller weights to interior regions, as illustrated in Fig.~\ref{fig:ode_high_causal}(a).
With this design, the high-frequency oscillation in the middle region can be partially recovered, as shown in Fig.~\ref{fig:ode_high}(c).
However, the solution near the two boundary regions is still not correctly fitted.
Moreover, because the boundary information cannot effectively propagate across the intermediate region to reach a globally consistent solution, a large low-frequency drift is eventually produced.
The loss curves in Fig.~\ref{fig:ode_high}(d) show that, although all loss terms can be optimized to some extent, they still cannot decrease to sufficiently low levels, indicating that the training becomes trapped in a poor local minimum.

Therefore, it is necessary to introduce the information bridge proposed in our method.
Specifically, we use a second-order polynomial to fit the predicted values at a set of anchor points during training, thereby constructing a low-frequency surrogate and generating pseudo-labels (see Fig.~\ref{fig:ode_high_causal}(b)).
This surrogate is then incorporated as a regularization term to pull back the low-frequency drift.
The resulting solution, shown in Fig.~\ref{fig:ode_high}(e), achieves high fitting accuracy.
The loss curves in Fig.~\ref{fig:ode_high}(f) further confirm that all loss components can be optimized to satisfactory levels.

\subsection{2D Poisson Equation with High Frequency Centers}

This case considers a standard two-dimensional Poisson equation, which is a representative diffusion-dominated problem.
Starting from an original globally smooth low-frequency source term, we further introduce four localized high-frequency components to increase the complexity of the solution structure.
Each high-frequency component is confined to a limited spatial region through a Gaussian window function, so that the resulting source term exhibits both global low-frequency variation and local high-frequency perturbations.
The solution near these local centers has a relatively large magnitude and varies rapidly, which makes the PDE operator locally stiff in these regions.
Therefore, this case provides a suitable benchmark for testing the proposed method, particularly its ability to handle locally stiff regions while preserving the overall solution structure.

The computational domain is partitioned into a \(5 \times 5\) grid of subregions, which are further grouped into three layers from the boundary toward the interior.
The BC weight is set to \(10{,}000\).
The quantitative results are reported in Table~\ref{tab:poissonhf_results}.
The standard PINN exhibits large prediction errors, and its PDE loss remains difficult to optimize, resulting in convergence to an incorrect solution.
In contrast, PINN-C1 effectively alleviates the training difficulty and captures the main solution structure.
The information bridge further suppresses low-frequency drift and improves global consistency, leading to a clear improvement in prediction accuracy.
With additional local region reweighting, PINN-C2 achieves a modest further improvement in convergence accuracy.

\begin{table}[htbp]
\centering
\caption{Quantitative results for the 2D Poisson Equation with High Frequency Centers.}
\label{tab:poissonhf_results}
\resizebox{\linewidth}{!}{%
\begin{tabular}{|l|c|c|c|c|}
\hline
\text{Method} & \text{PDE Residual} & $L_2$ Relative Error & \text{Max Abs Error} & \text{MSE} \\
\hline
PINN
& $(1.071 \pm 0.0001)\times 10^{3}$
& $(7.85 \pm 0.01)\times 10^{-1}$
& $(4.89 \pm 0.01)\times 10^{0}$
& $(3.98 \pm 0.01)\times 10^{-1}$ \\
\hline
PINN-C1(w/o bridge)
& $(9.49 \pm 3.65)\times 10^{0}$
& $(1.40 \pm 1.84)\times 10^{-1}$
& $(2.67 \pm 2.77)\times 10^{-1}$
& $(3.43 \pm 4.84)\times 10^{-2}$ \\
\hline
PINN-C1(with bridge)
& $(7.70 \pm 0.99)\times 10^{0}$
& $(1.78 \pm 0.87)\times 10^{-2}$
& $(5.41 \pm 0.03)\times 10^{-2}$
& $(2.53 \pm 2.25)\times 10^{-4}$ \\
\hline
PINN-C2
& $\mathbf{(7.41 \pm 1.00)\times 10^{0}}$
& $\mathbf{(1.05 \pm 0.35)\times 10^{-2}}$
& $\mathbf{(5.16 \pm 0.27)\times 10^{-2}}$
& $\mathbf{(7.93 \pm 5.32)\times 10^{-5}}$ \\
\hline
\end{tabular}%
}
\end{table}

\begin{figure}[!htbp]
\centering
  \subfigure[]{\includegraphics[width=0.7\linewidth]{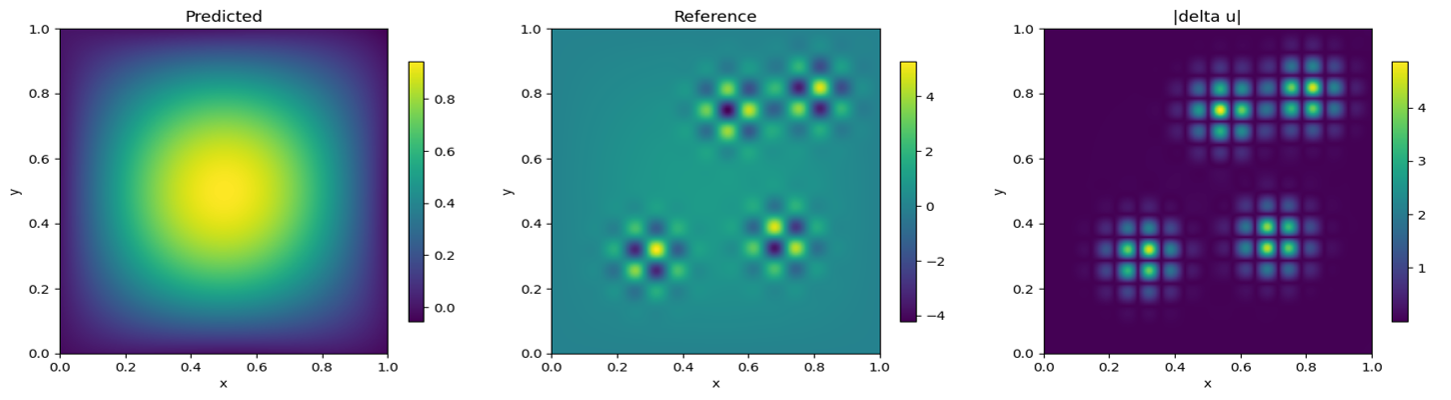}}\hspace{0.05\linewidth}
  \subfigure[]{\includegraphics[width=0.2\linewidth]{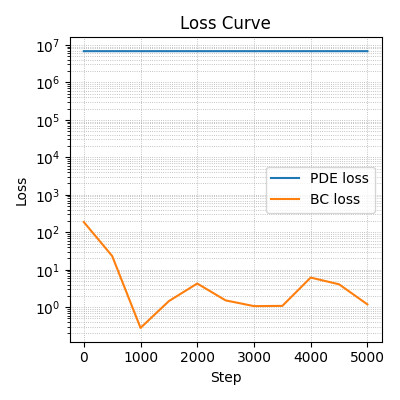}}
  \subfigure[]{\includegraphics[width=0.7\linewidth]{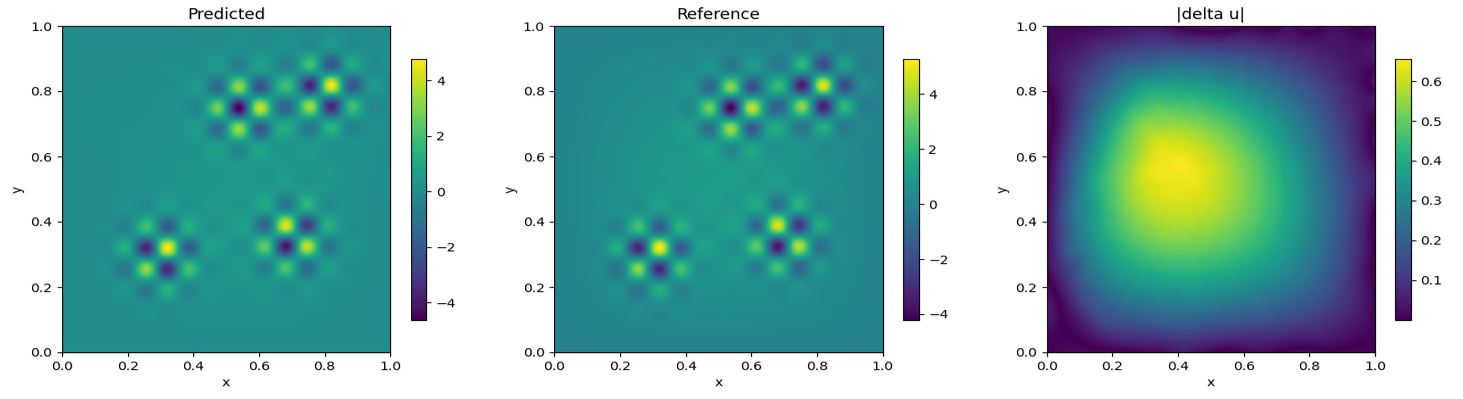}}\hspace{0.05\linewidth}
  \subfigure[]{\includegraphics[width=0.2\linewidth]{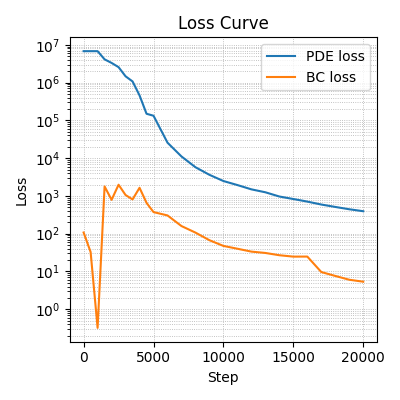}}
  \subfigure[]{\includegraphics[width=0.7\linewidth]{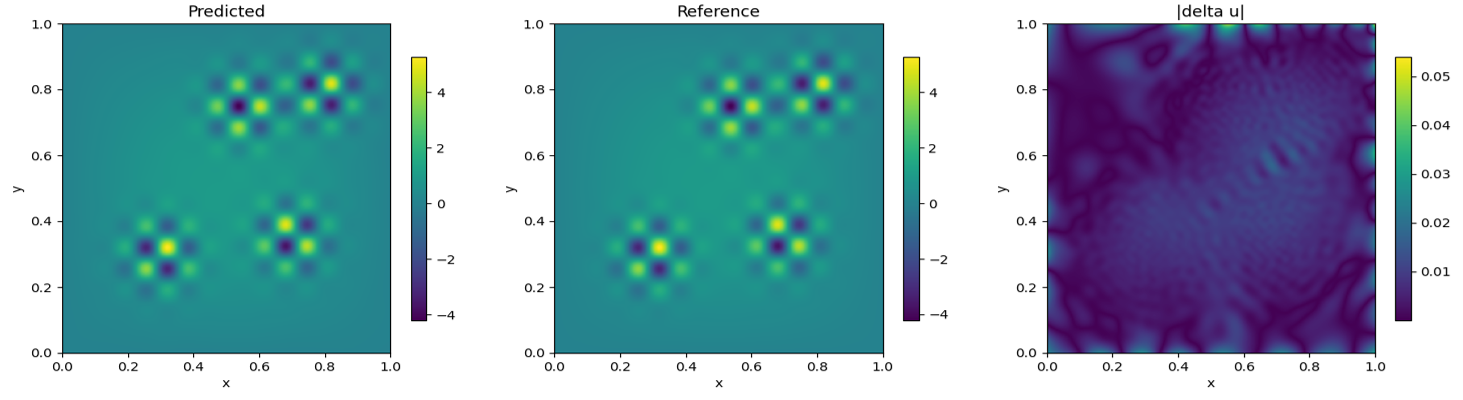}}\hspace{0.05\linewidth}
  \subfigure[]{\includegraphics[width=0.2\linewidth]{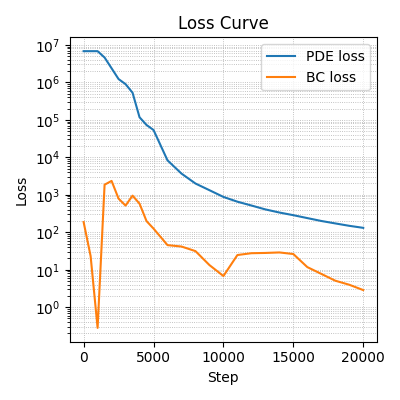}}
  \caption{\textbf{Results for the PoissonHF case.}
          (a)(b) The classical PINN captures only the low-frequency component and fails to resolve localized high-frequency details.
          (c)(d) PINN-C1(w/o bridge) recovers the main solution structure but exhibits low-frequency drift.
          (e)(f) PINN-C2 accurately recovers both the global structure and local details, with stable convergence of the PDE and BC losses.}
  \label{fig:poissonhf}
\end{figure}

Figure~\ref{fig:poissonhf} provides a visual comparison of the solution results.
As shown in Fig.~\ref{fig:poissonhf}(a), the classical PINN mainly captures the low-frequency component of the solution, while the PDE loss associated with the localized high-frequency structures remains difficult to reduce.
This behavior is also reflected in the loss curves in Fig.~\ref{fig:poissonhf}(b), where the second-stage L-BFGS optimization terminates early due to poor convergence.
As a result, the predicted solution fails to recover the localized high-frequency details.

In contrast, PINN-C1 without the information bridge can better fit the high-frequency structures.
However, it may still suffer from low-frequency drift, leading to an inaccurate global solution profile, as shown in Fig.~\ref{fig:poissonhf}(c)(d).
After incorporating the information bridge and region-adaptive  reweighting, the proposed PINN-C2, shown in Fig.~\ref{fig:poissonhf}(e), not only recovers the global solution structure but also preserves the local high-frequency details more accurately.
Throughout the optimization process, except for the initial warm-up stage, both the PDE loss and the BC loss decrease steadily to low levels, as shown in Fig.~\ref{fig:poissonhf}(f).

\begin{figure}[!htbp]
\centering
  \subfigure[]{\includegraphics[width=0.4\linewidth]{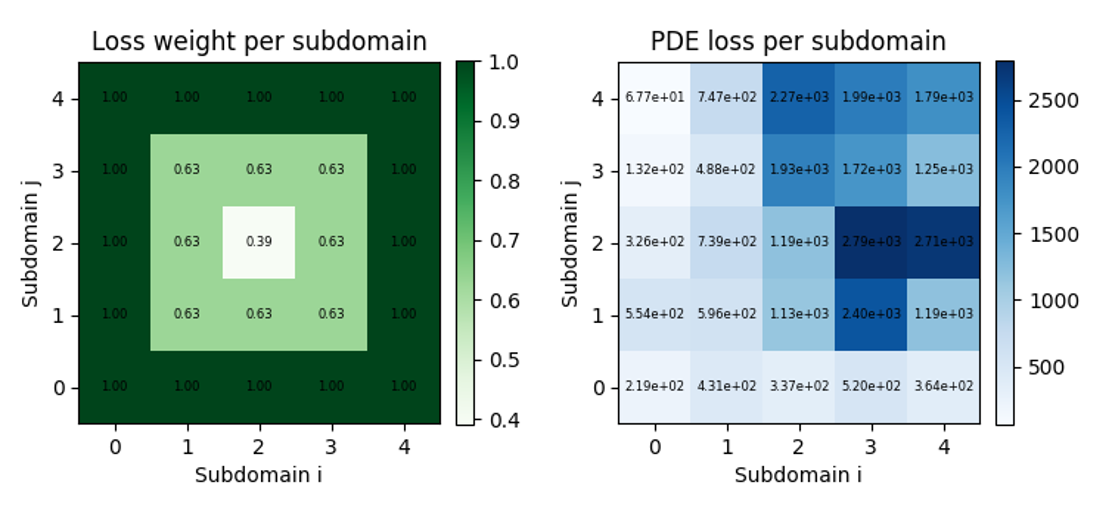}} \hspace{0.05\linewidth}
  \subfigure[]{\includegraphics[width=0.22\linewidth]{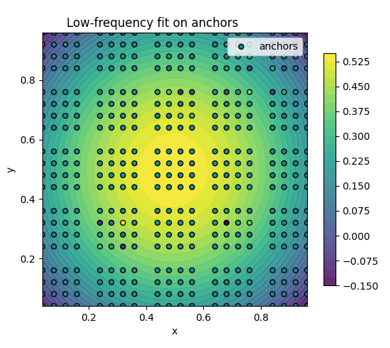}}
  \subfigure[]{\includegraphics[width=0.6\linewidth]{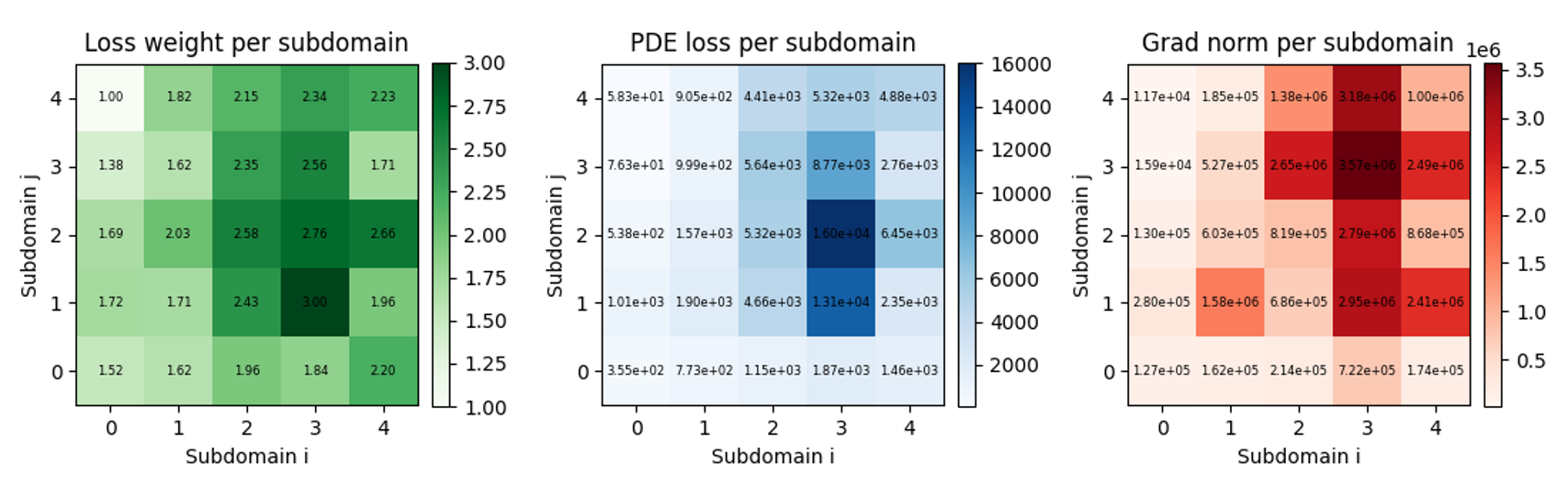}}
  \caption{\textbf{Intermediate training process for the PoissonHF case.}
      (a) Phase-I causal weighting from the boundary toward the interior.
      (b) Information bridge constructed from anchor points using a two-dimensional quadratic polynomial.
      (c) Phase-II adaptive local reweighting based on regional PDE losses and gradient norms.}
  \label{fig:poissonhf_weight}
\end{figure}

The intermediate training snapshots of the proposed method are shown in Fig.~\ref{fig:poissonhf_weight}.
In Phase I, the curriculum strategy guides information propagation from the boundary toward the interior.
Accordingly, the outer-layer regions are assigned larger weights, even though some interior subregions may exhibit relatively large PDE losses, as illustrated in Fig.~\ref{fig:poissonhf_weight}(a).
Figure~\ref{fig:poissonhf_weight}(b) shows the result of the information bridge, where a quadratic two-dimensional polynomial is fitted to the anchor points.
In Phase II, the local weights are further adjusted according to the PDE loss levels and gradient norms in different regions, allowing the training process to focus more effectively on high-error regions that receive insufficient gradient updates, as shown in Fig.~\ref{fig:poissonhf_weight}(c).

\subsection{2D Advection-Diffusion-Reaction Equation}

This case considers a two-dimensional advection-diffusion-reaction (ADR) equation.
Compared with purely diffusion-dominated problems, the ADR system exhibits richer physical behavior and more challenging training dynamics, since the solution is jointly influenced by directional advection, local smoothing, and spatially varying reaction effects.
From the training perspective, the coexistence of advection, diffusion, and reaction makes the PDE residual more difficult to optimize than in simpler elliptic cases.

In particular, the advection term introduces a preferred transport direction for information propagation, while the diffusion term tends to smooth the solution and the reaction term further modifies the local amplitude and stiffness of the operator.
Therefore, this case serves as a useful benchmark for testing whether the proposed spatial curriculum learning framework can guide information propagation more effectively and improve the fitting of locally difficult regions.

The computational domain is divided into a \(5 \times 5\) array of subregions and further organized into three layers from the boundary toward the interior.
The BC weight is fixed at 10,000, and the quantitative results are summarized in Table~\ref{tab:adr_results}.

Although this ADR case involves multiple coupled physical processes, the analytical solution is not particularly complex, as some terms are partially canceled by construction.
As a result, this case does not exhibit the severe loss stagnation, training failure, or low-frequency drift observed in more challenging examples.
Even in this relatively moderate setting, the quantitative results still demonstrate the overall effectiveness of the proposed framework.
Compared with the standard PINN, all proposed variants achieve a clear reduction in the PDE residual.
Among them, PINN-C1(with bridge) obtains the lowest PDE residual, whereas PINN-C2 achieves the best performance across all prediction error metrics.
These results suggest that the proposed framework can still provide beneficial effects for PDE problems involving multiple interacting physical mechanisms.

\begin{table}[htbp]
\centering
\caption{Quantitative results for the 2D Advection-Diffusion-Reaction Equation.}
\label{tab:adr_results}
\resizebox{\linewidth}{!}{%
\begin{tabular}{|l|c|c|c|c|}
\hline
\text{Method} & \text{PDE Residual} & $L_2$ Relative Error & \text{Max Abs Error} & \text{MSE} \\
\hline
PINN
& $(4.894 \pm 0.483)\times 10^{-1}$
& $(7.82 \pm 0.73)\times 10^{-3}$
& $(2.48 \pm 0.41)\times 10^{-2}$
& $(1.53 \pm 0.29)\times 10^{-5}$ \\
\hline
PINN-C1(w/o bridge)
& $(3.480 \pm 0.521)\times 10^{-1}$
& $(6.70 \pm 1.70)\times 10^{-3}$
& $(2.22 \pm 1.01)\times 10^{-2}$
& $(1.18 \pm 0.59)\times 10^{-5}$ \\
\hline
PINN-C1(with bridge)
& $\mathbf{(3.477 \pm 0.835)\times 10^{-1}}$
& $(6.32 \pm 2.35)\times 10^{-3}$
& $(2.58 \pm 1.47)\times 10^{-2}$
& $(1.12 \pm 0.83)\times 10^{-5}$ \\
\hline
PINN-C2
& $(3.495 \pm 0.364)\times 10^{-1}$
& $\mathbf{(5.71 \pm 0.27)\times 10^{-3}}$
& $\mathbf{(1.56 \pm 0.17)\times 10^{-2}}$
& $\mathbf{(8.08 \pm 0.75)\times 10^{-6}}$ \\
\hline
\end{tabular}%
}
\end{table}

\begin{figure}[!htbp]
\centering
  \subfigure[]{
    \includegraphics[width=0.7\linewidth]{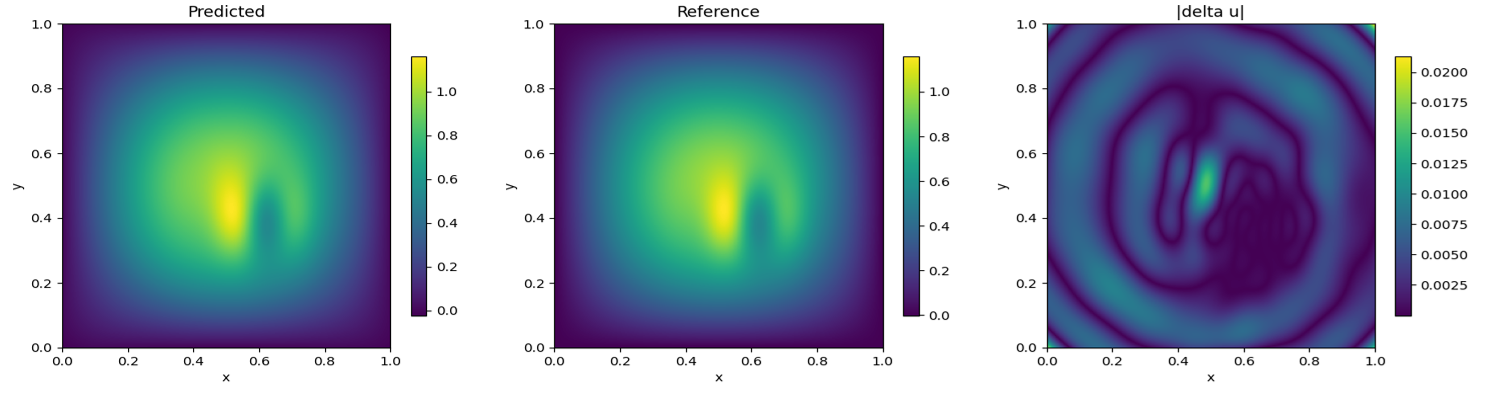}
  }\hspace{0.05\linewidth}
  \subfigure[]{
    \includegraphics[width=0.2\linewidth]{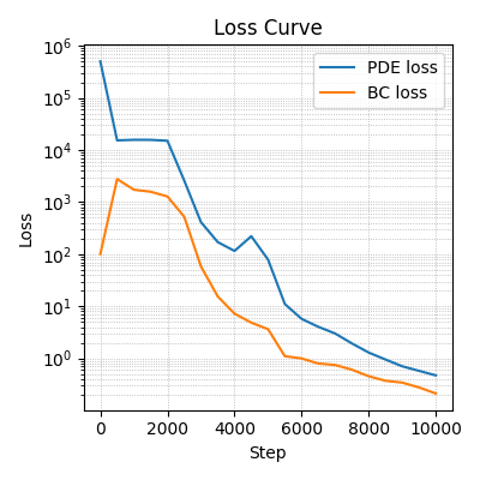}
  }

  \vspace{0.5em}

  \subfigure[]{
    \includegraphics[width=0.7\linewidth]{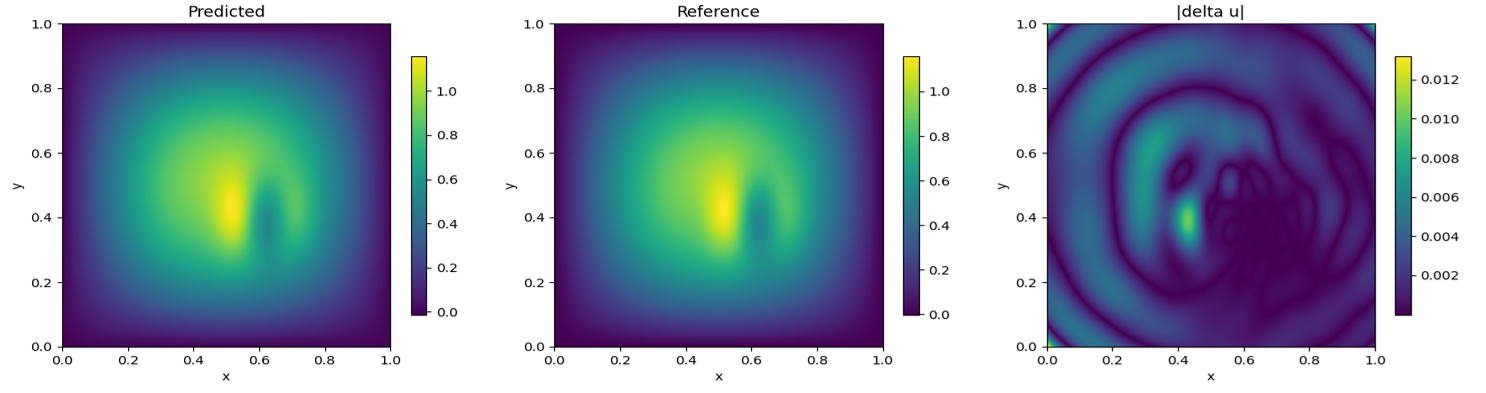}
  }\hspace{0.05\linewidth}
  \subfigure[]{
    \includegraphics[width=0.2\linewidth]{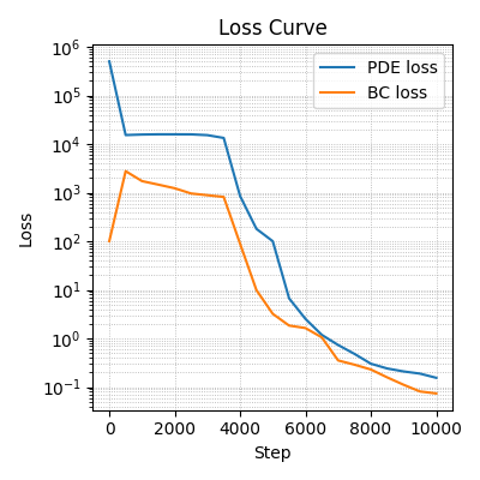}
  }

  \caption{\textbf{Results for the 2D advection--diffusion--reaction equation.}
  (a)(b) Classical PINN, which shows limited PDE-loss convergence and relatively large prediction errors:
  $\mathrm{MSE}=1.9192\times 10^{-5}$, $L_2$ relative error $=8.8126\times 10^{-3}$, max absolute error $=2.1332\times 10^{-2}$, and PDE residual $=5.2266\times 10^{-1}$.
  (c)(d) PINN-C2, which improves both solution accuracy and convergence behavior:
  $\mathrm{MSE}=7.0912\times 10^{-6}$, $L_2$ relative error $=5.3567\times 10^{-3}$, max absolute error $=1.3212\times 10^{-2}$, and PDE residual $=2.9840\times 10^{-1}$.}
  \label{fig:adr}
\end{figure}

Figure~\ref{fig:adr} shows the visual comparison of the solution results for the 2D advection-diffusion-reaction equation.
Unlike the previous cases, this problem does not suffer from severe training failure; the main difference between the methods is instead reflected in their final prediction accuracy.
As shown in Fig.~\ref{fig:adr}(a), the classical PINN already captures the overall solution structure, but its accuracy is still limited, especially in local regions.
For ease of comparison, the quantitative metrics are included directly in the figure caption.
Overall, PINN-C2 achieves better quantitative performance, including lower MSE, relative \(L_2\) error, maximum absolute error, and PDE residual.
As shown in Fig.~\ref{fig:adr}(c), PINN-C2 further improves the solution accuracy and local detail reconstruction.
As shown in Fig.~\ref{fig:adr}(d), the proposed method achieves a lower final loss than the baseline PINN, indicating more effective optimization.

\begin{figure}[!htbp]
\centering
  \subfigure[]{\includegraphics[width=0.4\linewidth]{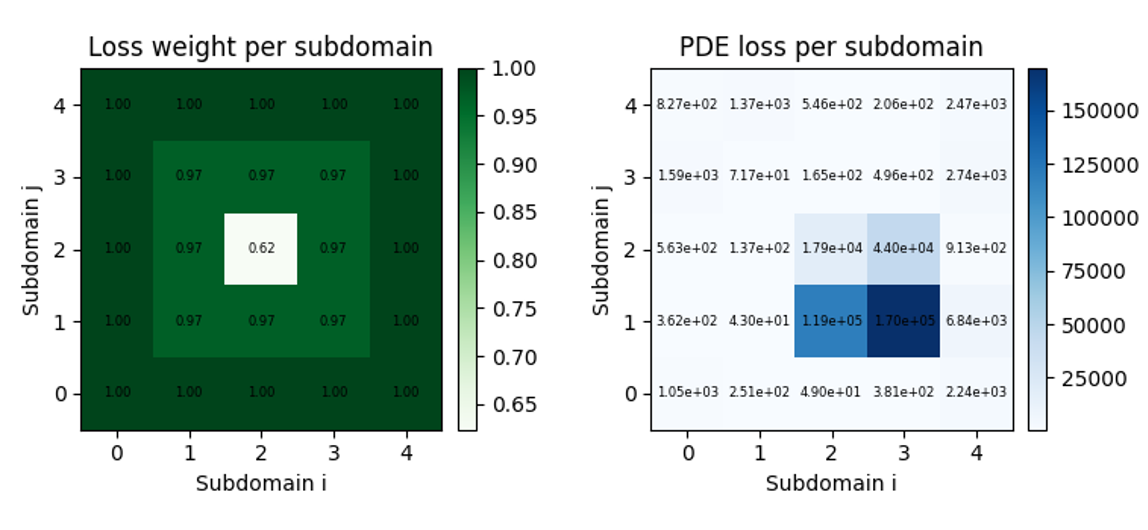}} \hspace{0.05\linewidth}
  \subfigure[]{\includegraphics[width=0.22\linewidth]{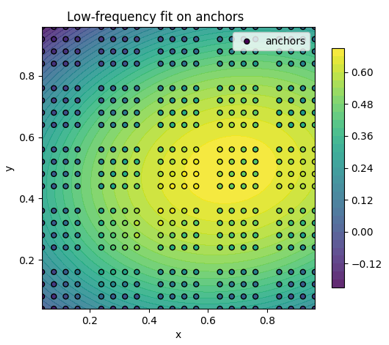}}
  \subfigure[]{\includegraphics[width=0.6\linewidth]{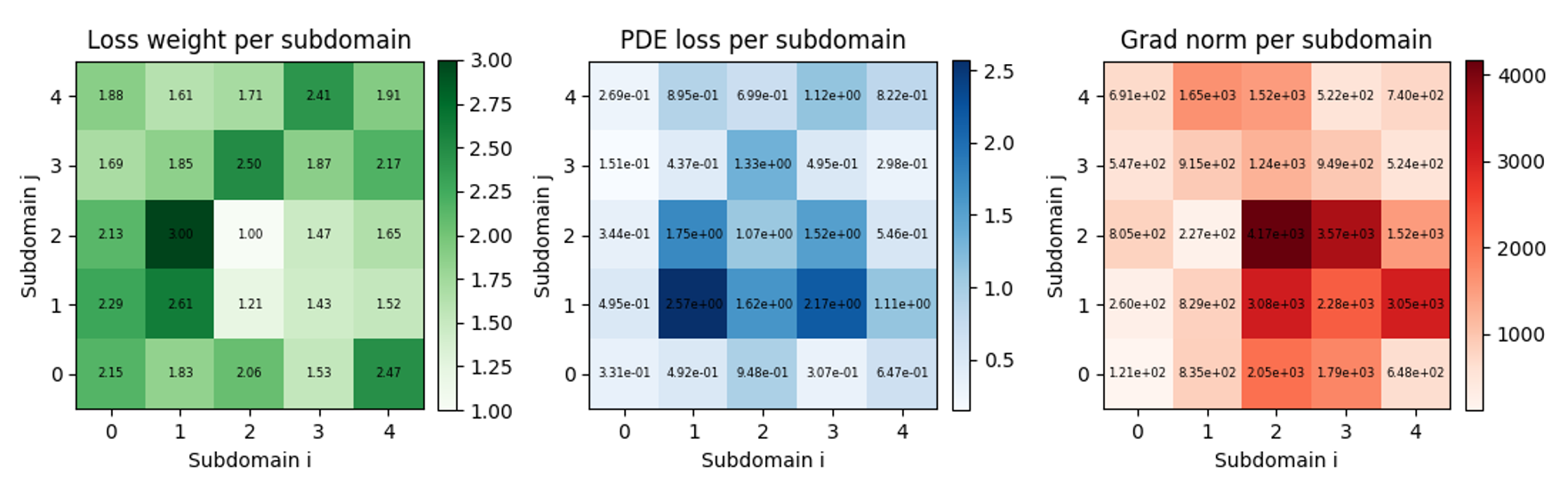}}
  \caption{\textbf{Intermediate training snapshots for the 2D advection--diffusion--reaction equation.}
    (a) Phase-I curriculum weighting, with larger weights assigned to the outer and middle layers than to the inner layer.
    (b) Information bridge constructed from anchor points using a two-dimensional quadratic polynomial.
    (c) Phase-II region-adaptive reweighting, which assigns a larger local weight to the lower-left region with large PDE loss and weak gradient influence.}
  \label{fig:adr_weight}
\end{figure}

Figure~\ref{fig:adr_weight} shows several intermediate snapshots of the proposed training process.
In Phase I, the curriculum strategy progressively propagates information from the boundary toward the interior.
At the stage shown in Fig.~\ref{fig:adr_weight}(a), the outermost layer already has a relatively small PDE loss, allowing the middle layer to receive substantial emphasis with a weight of 0.97.
By contrast, because some middle-layer subregions still have relatively large PDE losses, the innermost layer is assigned a smaller weight of 0.62 and is not yet strongly emphasized.
Figure~\ref{fig:adr_weight}(b) illustrates the information bridge constructed by fitting a quadratic two-dimensional polynomial to the anchor points, which captures the low-frequency diffusion trend well.
In Phase II, the local weights are further adapted according to the regional PDE losses and gradient norms.
As shown in Fig.~\ref{fig:adr_weight}(c), the lower-left region exhibits a larger PDE loss but a smaller gradient norm, indicating insufficient effective updates during training.
It is therefore assigned a larger local weight, so that the optimization process can focus more effectively on this difficult region.

\subsection{NS Equations for Lid-Driven Cavity Flow}

This case considers the incompressible Navier--Stokes equations for the classical lid-driven cavity flow, a standard benchmark in computational fluid dynamics.
The flow is defined in a closed square cavity, where the top boundary moves horizontally with a prescribed velocity profile depending on the horizontal coordinate, while the other three walls satisfy no-slip conditions.
A pressure reference point is specified at the lower-left corner to fix the pressure.
Driven by the moving lid, the fluid develops a recirculating flow pattern with a primary central vortex and possible secondary vortices near the corners.

Physically, this problem reflects the interaction between advection and viscous diffusion in a confined domain.
The Reynolds number is set to 100. The BC loss weights are set to \(100\).
Therefore, this case provides a suitable benchmark for evaluating whether the proposed framework can capture both the global flow structure and the localized high-gradient regions in a coupled nonlinear PDE system.

The quantitative results are reported in Tables~\ref{tab:ns_pressure_results} and~\ref{tab:ns_velocity_results}.
Compared with the standard PINN, PINN-C1 brings only limited improvement, and the use of causal weights alone even slightly degrades some metrics.
However, PINN-C1(with bridge) performs better than PINN-C1(w/o bridge), showing that the information bridge helps improve solution consistency.

With the Phase-II region-adaptive reweighting, PINN-C2 achieves the best performance in all metrics except the maximum absolute error of pressure.
The improvement is especially evident for the velocity components, indicating that local adaptive reweighting plays a more important role in this case.
This suggests that, even without severe training stagnation or obvious optimization failure, the proposed framework can still improve the final converged accuracy by enhancing local detail reconstruction and global solution quality.

\begin{table}[htbp]
\centering
\caption{Quantitative results for the NS equations: PDE residual and pressure-related metrics.}
\label{tab:ns_pressure_results}
\resizebox{\linewidth}{!}{%
\begin{tabular}{|l|c|c|c|c|}
\hline
\text{Method} & \text{PDE Residual} & $p$ $L_2$ Relative Error & $p$ Max Abs Error & $p$ MSE \\
\hline
PINN
& $(3.03 \pm 0.24)\times 10^{-2}$
& $(9.83 \pm 1.30)\times 10^{-2}$
& $(3.84 \pm 0.14)\times 10^{-1}$
& $(4.55 \pm 1.14)\times 10^{-4}$ \\
\hline
PINN-C1(w/o bridge)
& $(3.18 \pm 0.44)\times 10^{-2}$
& $(1.13 \pm 0.25)\times 10^{-1}$
& $\mathbf{(3.81 \pm 0.19)\times 10^{-1}}$
& $(6.20 \pm 2.55)\times 10^{-4}$ \\
\hline
PINN-C1(with bridge)
& $(3.03 \pm 0.44)\times 10^{-2}$
& $(9.05 \pm 1.68)\times 10^{-2}$
& $(3.82 \pm 0.20)\times 10^{-1}$
& $(3.93 \pm 1.50)\times 10^{-4}$ \\
\hline
PINN-C2
& $\mathbf{(2.12 \pm 0.35)\times 10^{-2}}$
& $\mathbf{(7.42 \pm 0.71)\times 10^{-2}}$
& $(3.95 \pm 0.22)\times 10^{-1}$
& $\mathbf{(2.57 \pm 0.50)\times 10^{-4}}$ \\
\hline
\end{tabular}%
}
\end{table}

\begin{table}[htbp]
\centering
\caption{Quantitative results for the NS equations: velocity-related metrics.}
\label{tab:ns_velocity_results}
\resizebox{\linewidth}{!}{%
\begin{tabular}{|l|c|c|c|c|c|c|}
\hline
\text{Method} 
& $u$ $L_2$ Relative Error 
& $u$ Max Abs Error 
& $u$ MSE 
& $v$ $L_2$ Relative Error 
& $v$ Max Abs Error 
& $v$ MSE \\
\hline
PINN
& $(7.09 \pm 1.41)\times 10^{-2}$
& $(8.17 \pm 2.01)\times 10^{-2}$
& $(7.45 \pm 2.85)\times 10^{-4}$
& $(1.06 \pm 0.15)\times 10^{-1}$
& $(1.02 \pm 0.15)\times 10^{-1}$
& $(7.83 \pm 2.24)\times 10^{-4}$ \\
\hline
PINN-C1(w/o bridge)
& $(9.02 \pm 2.46)\times 10^{-2}$
& $(1.02 \pm 0.29)\times 10^{-1}$
& $(1.24 \pm 0.62)\times 10^{-3}$
& $(1.33 \pm 0.35)\times 10^{-1}$
& $(1.29 \pm 0.37)\times 10^{-1}$
& $(1.30 \pm 0.62)\times 10^{-3}$ \\
\hline
PINN-C1(with bridge)
& $(6.44 \pm 1.61)\times 10^{-2}$
& $(7.40 \pm 1.86)\times 10^{-2}$
& $(6.28 \pm 3.21)\times 10^{-4}$
& $(9.66 \pm 2.14)\times 10^{-2}$
& $(9.38 \pm 2.10)\times 10^{-2}$
& $(6.74 \pm 3.07)\times 10^{-4}$ \\
\hline
PINN-C2
& $\mathbf{(4.55 \pm 0.47)\times 10^{-2}}$
& $\mathbf{(5.33 \pm 0.74)\times 10^{-2}}$
& $\mathbf{(2.98 \pm 0.63)\times 10^{-4}}$
& $\mathbf{(6.66 \pm 0.65)\times 10^{-2}}$
& $\mathbf{(5.92 \pm 0.85)\times 10^{-2}}$
& $\mathbf{(3.08 \pm 0.61)\times 10^{-4}}$ \\
\hline
\end{tabular}%
}
\end{table}

\begin{figure}[!htbp]
\centering
  \subfigure[]{\includegraphics[width=0.173\linewidth]{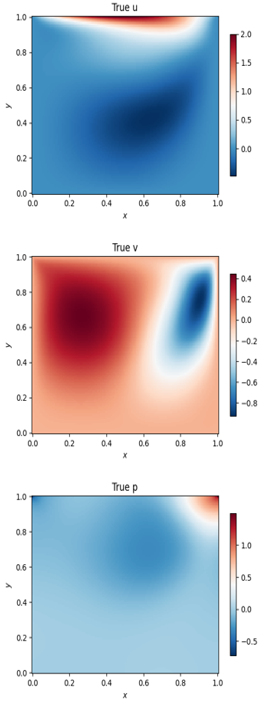}}\hspace{0.05\linewidth}
  \subfigure[]{\includegraphics[width=0.35\linewidth]{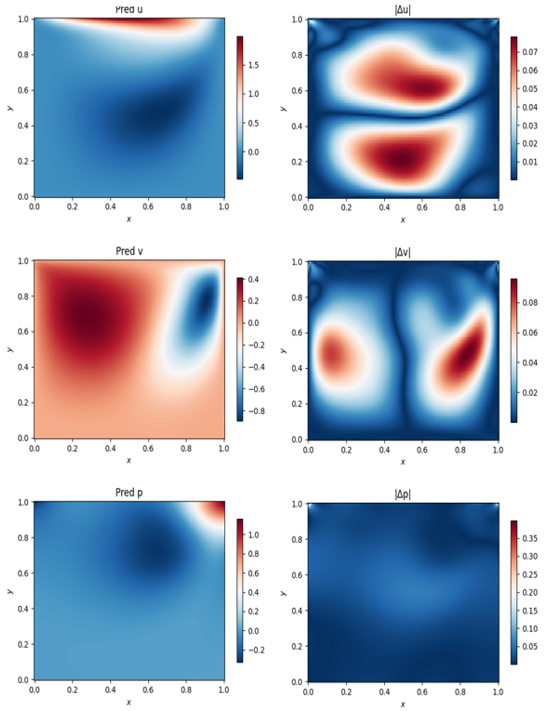}}\hspace{0.05\linewidth}
  \subfigure[]{\includegraphics[width=0.35\linewidth]{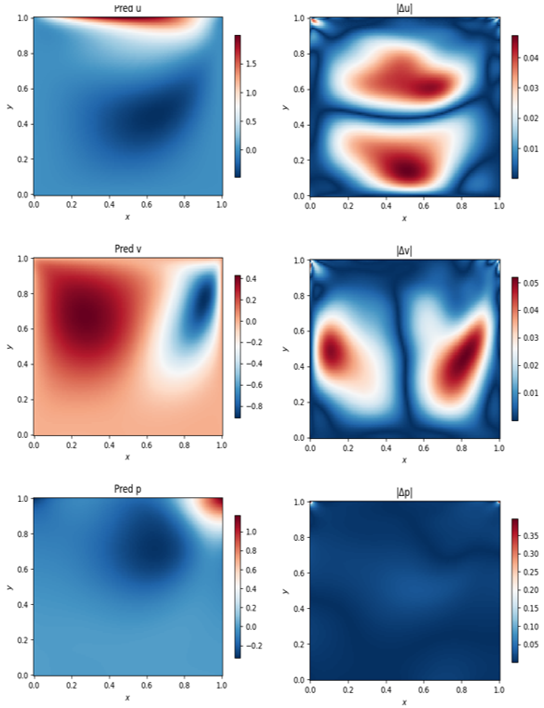}}
  \caption{\textbf{Prediction results for the incompressible Navier--Stokes equations in the lid-driven cavity flow case.}
  (a) Reference solution.
  (b) Prediction results and absolute error distribution of the baseline PINN, which gives a PDE residual of \(3.1058\times 10^{-2}\) and \(L_2\) relative errors of \(1.0454\times 10^{-1}\), \(7.0820\times 10^{-2}\), and \(1.0492\times 10^{-1}\) for \(p\), \(u\), and \(v\), respectively.
  (c) Prediction results and absolute error distribution of PINN-C2, which reduces the PDE residual to \(1.9113\times 10^{-2}\) and the corresponding \(L_2\) relative errors to \(6.7516\times 10^{-2}\), \(4.2435\times 10^{-2}\), and \(6.2153\times 10^{-2}\).}
  \label{fig:ns}
\end{figure}

As can also be observed from the visualizations in Fig.~\ref{fig:ns}, the prediction results of the baseline PINN and PINN-C2 are generally close at the global level.
Nevertheless, PINN-C2 still shows noticeable improvements in both overall accuracy and local detail reconstruction, with the main error metrics reduced by approximately 40\%.

\begin{figure}[!htbp]
\centering
  \subfigure[]{\includegraphics[width=0.37\linewidth]{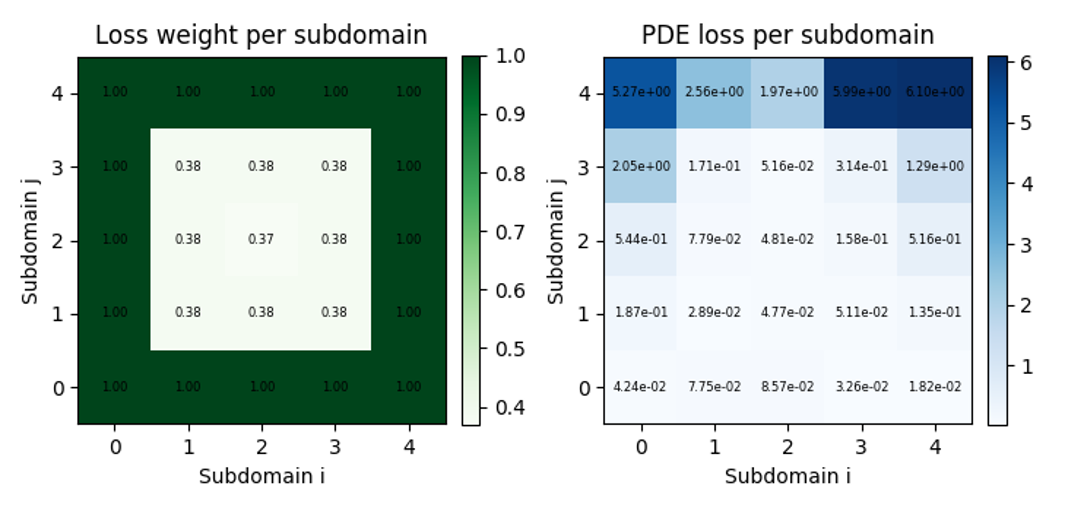}}\hspace{0.03\linewidth}
  \subfigure[]{\includegraphics[width=0.56\linewidth]{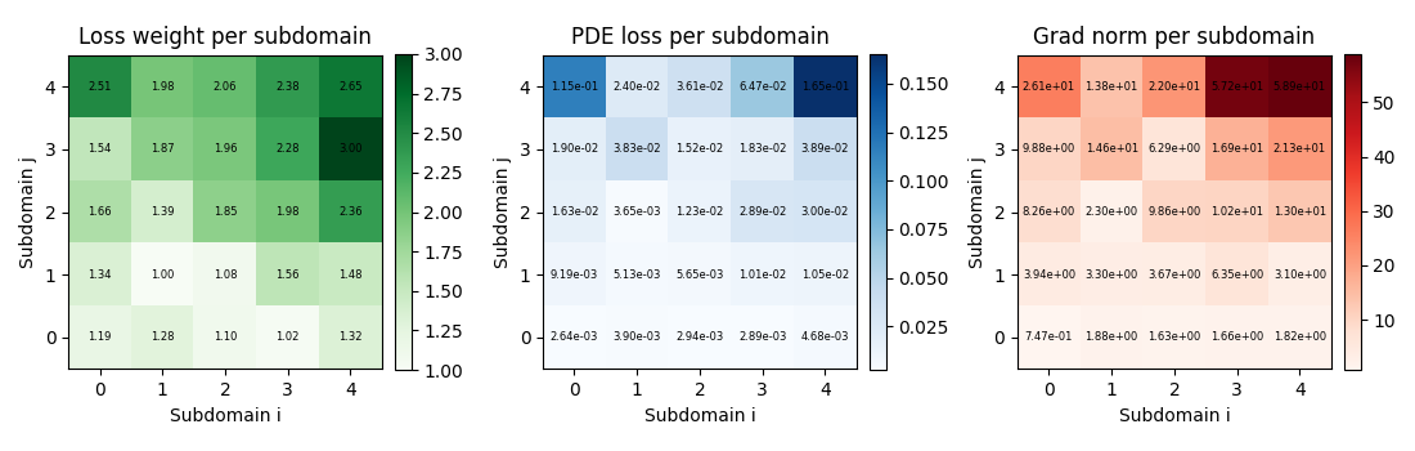}}
  \subfigure[]{\includegraphics[width=0.6\linewidth]{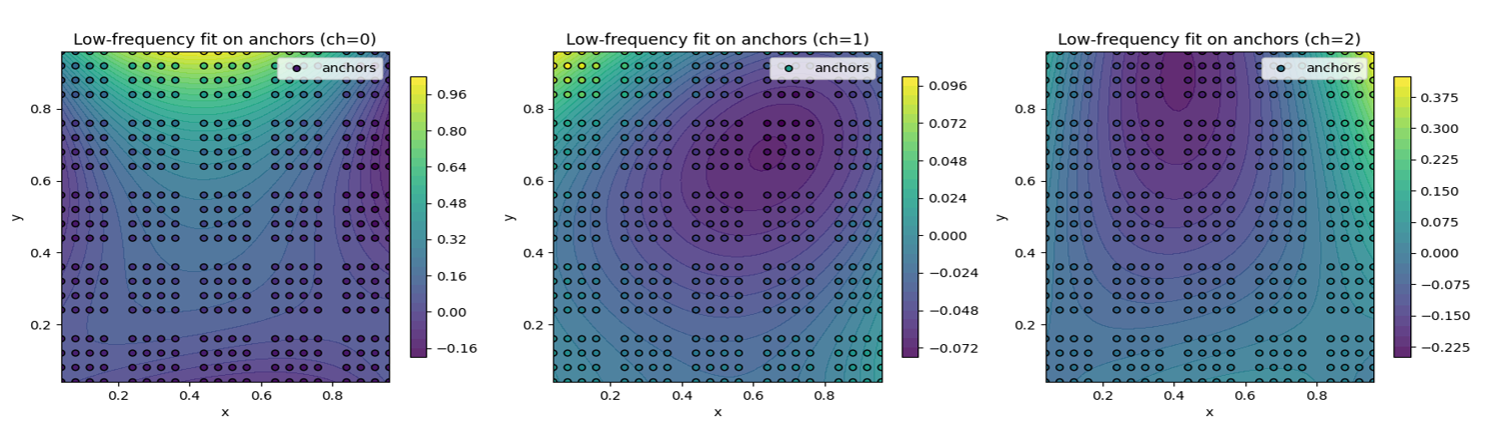}}
  \caption{\textbf{Visualization of the proposed training strategy for the incompressible Navier--Stokes equations in the lid-driven cavity flow case.}
  (a) Phase-I curriculum weighting, emphasizing the upper region and the upper-right corner to guide boundary-to-interior learning.
  (b) Phase-II adaptive local reweighting based on regional PDE losses and gradient norms, assigning the largest weight to the region slightly below the upper-right corner after their joint assessment.
  (c) Information bridge fitted from anchor points using a third-order two-dimensional polynomial, where \texttt{ch}=0, 1, and 2 correspond to the low-frequency structures of \(u\), \(v\), and \(p\), respectively.}
  \label{fig:ns_weight}
\end{figure}

Fig.~\ref{fig:ns_weight} illustrates the training process of the proposed curriculum framework. 
In Fig.~\ref{fig:ns_weight}(a), larger causal weights are assigned to the upper region and the upper-right corner, where the PDE loss has not yet fully converged. 
Even when the PDE loss in some interior regions is already small, the training is still guided to focus first on the outer layer, so as to establish a more reliable boundary-induced solution structure.
Fig.~\ref{fig:ns_weight}(b) shows the adaptive local reweighting in the second phase.
Although the upper-left and upper-right regions both exhibit relatively large PDE losses and gradient norms, the joint assessment assigns the largest adaptive weight to the region slightly below the upper-right corner.
This indicates that the proposed strategy emphasizes the region requiring stronger optimization based on both residual magnitude and gradient information.
Fig.~\ref{fig:ns_weight}(c) presents the information bridge fitted by a third-order polynomial, where the three channels (\texttt{ch}=0,1,2) correspond to the low-frequency structures of \(u\), \(v\), and \(p\), respectively. 
The first channel captures the relatively large horizontal velocity near the upper boundary. 
The second channel, limited by the low polynomial order, does not fully reproduce the asymmetric distribution of the \(v\)-component. 
The third channel reflects the low-frequency trend of the pressure field, with relatively large values near the upper-right corner.

\section{Conclusion}

This paper proposes a curriculum learning framework for PINNs from the perspective of spatial correlation, with a particular focus on coupled PDE boundary value problems.
Unlike existing approaches that mainly rely on temporal causality or simultaneous global optimization, the proposed method reformulates the curriculum mechanism of PINNs as a spatially progressive optimization process, consisting of boundary information propagation, cross-region consistency regularization, and gradual refinement of local details.

To this end, three key components are introduced.
First, spatial causal weights are introduced to impose a weak-causality constraint, guiding boundary information from near-boundary regions toward the interior and thereby reducing the difficulty of nonconvex optimization as well as the conflict between the boundary loss and the PDE residual loss.
Second, an information bridge mechanism is proposed to suppress low-frequency drift during cross-region training, thus improving the global consistency and stability of the learned solution.
Third, a region-adaptive weighting strategy is developed to further enhance the fitting of locally high-frequency structures and difficult regions while preserving the overall propagation pattern.
Experimental results demonstrate that the proposed framework can effectively improve the training behavior of PINNs across multiple cases, yielding clear gains in training stability, the effectiveness of information propagation, and final prediction accuracy.

This work also suggests several directions for future research.
Because different PDEs exhibit distinct physical characteristics and optimization challenges, more physically adaptive spatial curriculum strategies are needed for different problem types.
Moreover, information propagation in complex systems is often richer than a simple boundary-to-interior process, especially under strong advection, closed-loop flows, or multi-path coupling, calling for more general formulations of spatial weak causality.
Extending the framework to irregular geometries, multiple boundaries, more general boundary conditions such as Robin boundary conditions, and insufficient boundary information is also an important direction.
Future work will focus on these aspects to develop a more general and robust spatial curriculum learning approach for PINNs.

%%%% Acknowledgments %%%%%%%%
% \section*{Acknowledgments}
% The author would like to thank  ....

\section*{Appendix A: Analysis of PDE Problem Characteristics}

To discuss phenomena such as "information propagation" and "global coupling" in PINN training more clearly, this appendix summarizes the characteristic types of common PDEs from both mathematical and physical perspectives, and briefly explains the decisive role of different initial/boundary conditions on the solution\textsuperscript{\cite{evans2022partial}}.

\subsection*{A.1 PDE Types}
From a mathematical perspective, second-order linear PDEs are classified based on the principal symbol quadratic form.
The principal part consists of second-order derivatives, and depending on the definiteness or degeneracy of the corresponding quadratic form, PDEs are classified as elliptic, hyperbolic, or parabolic types\textsuperscript{\cite{doi:10.1142/9789813229181_0002}}.. These classifications influence the solution's coupling range, propagation mechanism, and characteristic direction.

\begin{itemize}
  \item \textbf{Elliptic:} Elliptic equations lack a temporal causal structure and clear propagation direction. Their solutions exhibit global spatial coupling, meaning local disturbances can affect the entire domain. They are common in steady-state diffusion and potential field problems, reflecting "global constraints without explicit propagation."
  
  \item \textbf{Hyperbolic:} Hyperbolic equations have strong causality, with solutions propagating along characteristic lines or surfaces. These equations are associated with wave propagation and strong convective transport, where information is transmitted along the characteristic direction, focusing on "propagation along the characteristics."
  
  \item \textbf{Parabolic:} Parabolic equations are "weakly causal," with solutions evolving in one direction, typically time, while diffusion smooths and weakens the information. The heat conduction equation is a typical example, where the solution advances over time and becomes increasingly diffused.
\end{itemize}

This classification reveals a key fact: the information propagation structure of different PDEs is not consistent. Therefore, training strategies centered on "causal progression" are often more natural for hyperbolic/parabolic equations, but extending this to elliptic equations or strongly coupled BVPs requires more careful design.

\subsection*{A.2 Physical Perspective}
From a physical perspective, many engineering PDEs can be represented within the Advection-Diffusion-Reaction (ADR) framework\textsuperscript{\cite{tariq2025unveiling}}, where each term corresponds to distinct "information dynamics," influencing both the spatial/temporal structure of the solution and the associated learning challenges.

\begin{itemize}
  \item \textbf{Transient vs. Steady-State:} Transient problems evolve over time, moving from an initial to a steady state. Steady-state problems, however, are not time-dependent, and their solutions are determined by spatial constraints and source terms. For PINNs, transient problems naturally introduce time propagation, while steady-state problems emphasize global consistency constraints.

  \item \textbf{Advection Term:} This term represents the "transport" effect, transferring information along the flow direction, typically showing clear directionality and asymmetry. In advection-dominated cases, such as with high Peclet numbers, the solution exhibits strong directional characteristics and boundary layers, leading to potential issues with "insufficient propagation along the characteristic direction" during training.

  \item \textbf{Diffusion Term:} This term represents "mixing" and "smoothing," weakening gradients and diffusing high-frequency structures. As diffusion increases, the solution smooths out, and local disturbances become less significant. 

  \item \textbf{Reaction Term:} Describes the local generation/consumption mechanism, affecting the growth or decay rate of the solution. When reaction dominates, the solution may localize or show rapid growth/decay, causing scale inconsistencies and leading to "local rigidity."
\end{itemize}

The three physical mechanisms of ADR are closely related to the mathematical classification of PDEs. 
In complex problems, such as the Navier-Stokes equations\textsuperscript{\cite{Doering_Gibbon_1995}}, they determine the solution's multi-scale nature, peak/boundary layer structures, and information propagation patterns. These mechanisms also serve as an essential physical foundation for understanding the challenges in PINN training.

\subsection*{A.3 Boundary Conditions and Problem Forms}
In addition to the equation type, the form of initial and boundary conditions directly determines the uniqueness of the solution and the mode of information input\textsuperscript{\cite{gilbarg1998elliptic}}:

\begin{itemize}
\item \textbf{Initial Value Problem (IVP):} A typical evolution problem, where the solution is typically uniquely determined by the initial conditions, provided the evolution direction aligns with the equation type (e.g., parabolic or hyperbolic equations advancing in time). In this case, information propagates primarily from the initial time onward.

\item \textbf{Boundary Value Problem (BVP) and Initial-Boundary Value Problem (IBVP):} Constraint problems that require sufficient boundary conditions on spatial boundaries (and possibly initial surfaces) to ensure the problem is well-posed. For elliptic and strongly coupled systems, boundary conditions often define the global solution shape. In such cases, information does not simply "propagate in one direction"; instead, it reflects global consistency and coupling across regions.
\end{itemize}

\section*{Appendix B: Mathematical Formulation of the PDEs}

\subsection*{B.1 1D ODE}

We consider a second-order boundary value problem (BVP) for a 1D ordinary differential equation (ODE) with the following formulations:

\begin{equation}
\frac{\mathrm{d}^2 y}{\mathrm{d}x^2} = e^{\alpha x} \left[ g_2(x) \sin(kx) + 2k g_1(x) \cos(kx) \right], \quad x \in (0, 1),
\end{equation}

where \(g_1(x)\) and \(g_2(x)\) are given by:

\begin{equation}
g_1(x) = 1 - x - x^2 + \alpha (x - x^2),
\end{equation}

\begin{equation}
g_2(x) = -2 + 2\alpha (1 - 2x) + (\alpha^2 - k^2)(x - x^2),
\end{equation}

with the boundary conditions:

\begin{equation}
y(0) = 0, \quad y(1) = 0.
\end{equation}

\paragraph{Parameter Settings}
To investigate the effect of spatial frequency and stiffness, two configurations are considered:

\[
\text{Low-frequency case:} \quad k = 10\pi, \quad \alpha = 2.0,
\]

\[
\text{High-frequency case:} \quad k = 20\pi, \quad \alpha = 3.0.
\]

\paragraph{Exact Solution}
The analytical exact solution to the ODE is given by:

\begin{equation}
y_{exact}(x) = x(1 - x) e^{\alpha x} \sin(k x)
\end{equation}

where \( \alpha \) and \( k \) are the parameters defined in the respective low-frequency and high-frequency cases.

\section*{B.2 2D Poisson Equation with High Frequency Centers}

Consider a 2D Poisson equation where the solution consists of a global low-frequency term and high-frequency components centered at four specific locations. The problem is defined as follows:

\begin{equation}
-\Delta u(x,y) = f(x,y), \qquad (x,y) \in \Omega,
\end{equation}

where the domain \(\Omega\) is:

\begin{equation}
\Omega = (0,1) \times (0,1),
\end{equation}

This problem uses Dirichlet boundary conditions, where the boundary values are given by the analytical solution:

\begin{equation}
u(x, y) = u_{\text{exact}}(x, y), \quad \forall (x, y) \in \partial \Omega.
\end{equation}

where \(u_{\text{exact}}(x,y)\) denotes the analytical solution and \(\partial \Omega\) denotes the boundary of the domain \(\Omega\).

\paragraph{Exact Solution}

This case is one we constructed, and we define its analytical solution \(u(x,y)\) as the sum of the low-frequency and high-frequency components:

\begin{equation}
u_{exact}(x,y) = u_0(x,y) + u_p(x,y),
\end{equation}

where the low-frequency part \(u_0(x,y)\) is:

\begin{equation}
u_0(x,y) = \sin(\pi x) \sin(\pi y),
\end{equation}

and the high-frequency part \(u_p(x,y)\) is:

\begin{equation}
u_p(x,y) = \sum_{m=1}^{M} a_m G_m(x,y) S(x,y), \qquad M=4,
\end{equation}

with the function \(S(x,y)\) defined as:

\begin{equation}
S(x,y) = \sin(k\pi x) \sin(k\pi y),
\end{equation}

and \(G_m(x,y)\) representing a Gaussian function centered at \((c_{x,m}, c_{y,m})\):

\begin{equation}
G_m(x,y) = \exp\left(-\frac{(x - c_{x,m})^2 + (y - c_{y,m})^2}{\sigma^2}\right).
\end{equation}

% \paragraph{Source Term Derivation}

% To derive the source term \( f(x, y) \), we need to compute the Laplacian of the solution \( u(x, y) \).

% \

% The Laplacian of the low-frequency component is

% \begin{equation}
% \Delta u_0(x,y) = -2\pi^2 \sin(\pi x) \sin(\pi y)
% \end{equation}

% The derivatives of \(S(x,y)\) and \(G_m(x,y)\) are

% \[
% S_x = k\pi \cos(k\pi x) \sin(k\pi y), \quad S_y = k\pi \sin(k\pi x) \cos(k\pi y),
% \]
% \[
% S_{xx} = -(k\pi)^2 \sin(k\pi x) \sin(k\pi y), \quad S_{yy} = -(k\pi)^2 \sin(k\pi x) \sin(k\pi y),
% \]
% \[
% G_{m,x} = -\frac{2(x - c_{x,m})}{\sigma^2} G_m, \quad G_{m,y} = -\frac{2(y - c_{y,m})}{\sigma^2} G_m,
% \]
% \[
% G_{m,xx} = \left(\frac{4(x - c_{x,m})^2}{\sigma^4} - \frac{2}{\sigma^2}\right) G_m, \quad G_{m,yy} = \left(\frac{4(y - c_{y,m})^2}{\sigma^4} - \frac{2}{\sigma^2}\right) G_m,
% \]

% \[
% \Delta \left(G_m S\right) = G_{m,xx} S + 2 G_{m,x} S_x + G_m S_{xx} + G_{m,yy} S + 2 G_{m,y} S_y + G_m S_{yy}.
% \]

% Therefore, the source term \(f(x,y)\) is

% \begin{equation}
% f(x,y) = -\Delta u(x,y) = 2\pi^2 \sin(\pi x) \sin(\pi y) - \sum_{m=1}^{M} a_m \, \Delta \left(G_m(x,y) S(x,y)\right)
% \end{equation}

\paragraph{Parameter Settings}
The parameters used in the 2D Poisson equation are as follows:

\

The frequency parameter is \( k \) and the width of the Gaussian functions is \( \sigma\), given by:

\[
k = 14, \quad \sigma = 0.1,
\]

The centers of the Gaussians are located at:

\[
\{(c_{x,m}, c_{y,m})\}_{m=1}^4 = \{(0.30, 0.30), (0.70, 0.35), (0.55, 0.75), (0.80, 0.80)\},
\]

The coefficients \(a_m\) are:

\[
(a_m)_{m=1}^4 = (5.0, 5.0, 5.0, 5.0).
\]

\section*{B.3 2D ADR Equation}

Consider the steady-state advection-diffusion-reaction (ADR) equation, which includes advection, reaction, and diffusion terms. The equation is given by:

\begin{equation}
-\nu \Delta u(x,y) + b(x,y) \cdot \nabla u(x,y) + c(x,y) \, u(x,y) = f(x,y), \qquad (x,y) \in \Omega,
\end{equation}

where the domain \(\Omega\) is:

\begin{equation}
\Omega = (0,1) \times (0,1),
\end{equation}

The advection term \(b(x,y)\) is defined as:

\begin{equation}
\boldsymbol{b}(x,y) = 
\begin{pmatrix}
b_x(x,y)\\[2pt]
b_y(x,y)
\end{pmatrix}
=
\beta
\begin{pmatrix}
-(y - 0.5)\\
x - 0.5
\end{pmatrix},
\end{equation}

The reaction term \(c(x,y)\) is:

\begin{equation}
c(x,y) = c_{\min} + c_{\max} \exp\left( -\frac{(x - x_c)^2 + (y - y_c)^2}{\sigma_c^2} \right),
\end{equation}

Dirichlet boundary conditions are given by:

\begin{equation}
u(x, y) = u_{\text{exact}}(x, y), \quad \forall (x, y) \in \partial \Omega.
\end{equation}

where \(u_{\text{exact}}(x,y)\) denotes the analytical solution and \(\partial \Omega\) denotes the boundary of the domain \(\Omega\).

\paragraph{Exact Solution}

The solution \(u(x,y)\) we constructed is expressed as the sum of the low-frequency part \(u_{\text{low}}(x,y)\) and a perturbation term \(\delta \cdot u_{\text{packet}}(x,y)\):

\begin{equation}
u_{exact}(x,y) = u_{\text{low}}(x,y) + \delta \cdot, u_{\text{packet}}(x,y),
\end{equation}

The low-frequency part \(u_{\text{low}}(x,y)\) is:

\begin{equation}
u_{\text{low}}(x,y) = \sin(\pi x) \sin(\pi y),
\end{equation}

The perturbation term \(u_{\text{packet}}(x,y)\) is:

\begin{equation}
u_{\text{packet}}(x,y) = \exp\left( -\frac{(x - x_0)^2 + (y - y_0)^2}{\sigma_u^2} \right) \cos(k\pi x),
\end{equation}

\paragraph{Parameter Settings}
The parameters used in the 2D ADR equation are as follows:

\[
\nu = 10^{-4}, \quad \beta = 100, \quad k = 8, \quad \delta = 0.35,
\]

\[
(x_0, y_0) = (0.60, 0.40), \quad (x_c, y_c) = (0.65, 0.35),
\]

\[
\sigma_u = 0.12, \quad \sigma_c = 0.10, \quad c_{\min} = 1, \quad c_{\max} = 10^{4}.
\]

\section*{B.4 2D NS Equations for Lid-Driven Cavity Flow}

Consider the steady-state Navier-Stokes (NS) equations for the 2D lid-driven cavity flow problem. There is no analytical solution for this problem, and the reference solution is obtained using COMSOL software. The problem is defined on the domain \(\Omega\) as:

\begin{equation}
\Omega = [0, 1] \times [0, 1]
\end{equation}

The steady-state Navier-Stokes equations are given by:

\begin{equation}
u u_x + v u_y + p_x - \frac{1}{\text{Re}}(u_{xx} + u_{yy}) = 0, \qquad (x, y) \in \Omega,
\end{equation}

\begin{equation}
u v_x + v v_y + p_y - \frac{1}{\text{Re}}(v_{xx} + v_{yy}) = 0, \qquad (x, y) \in \Omega,
\end{equation}

\begin{equation}
u_x + v_y = 0, \qquad (x, y) \in \Omega.
\end{equation}

The velocity boundary conditions are as follows:

\begin{equation}
u(x, 1) = a x (1 - x), \qquad v(x, 1) = 0, \qquad x \in [0, 1],
\end{equation}

\begin{equation}
u(x, y) = 0, \qquad v(x, y) = 0, \qquad (x, y) \in \partial \Omega \setminus \{y = 1\}.
\end{equation}

The pressure boundary condition is:

\begin{equation}
p(0, 0) = 0.
\end{equation}

\paragraph{Parameter Settings}

The Reynolds number and lid velocity coefficient are solved set as:

\[ \text{Re} = 100, \quad a = 8. \]

\section*{Appendix C: Experimental Parameter Settings and Details}

This section provides the detailed experimental parameters and configuration used throughout the experiments, including the setup for different cases, training parameters, and configurations for the reweighting strategy.

% \paragraph{Case Parameters}
\begin{table}[htbp] 
\centering
\caption{Parameter settings for Case B.1 under different frequency settings}
\label{tab:parameters_b1}
\begin{tabular}{|l|c|c|}
\hline
\textbf{Parameter} & \textbf{B.1 (low frequency)} & \textbf{B.1 (high frequency)} \\
\hline
Number of domain points & 250 & 1000 \\
\hline
Number of boundary points & 2 & 2 \\
\hline
Learning rate & \(1\times 10^{-3}\) & \(1\times 10^{-3}\) \\
\hline
Adam iterations & 10000 & 10000 \\
\hline
L-BFGS iterations & 0 & 6000 \\
\hline
Optimizer decay & 2000, 0.9 & 2000, 0.9 \\
\hline
Network width & 100 & 100 \\
\hline
Network depth & 6 & 6 \\
\hline
\end{tabular}
\end{table}

\begin{table}[htbp]
\centering
\caption{Reweighting configuration for Case B.1 under different frequency settings}
\label{tab:reweighting_b1}
\begin{tabular}{|l|c|c|}
\hline
\textbf{Parameter} & \textbf{B.1 (low frequency)} & \textbf{B.1 (high frequency)} \\
\hline
Decay epsilon & 1.0 & 1.0 \\
\hline
Number of subregions & 5 & 5 \\
\hline
Reweight every & 500 & 500 \\
\hline
Reweight causal steps & 1000 - 5000 & 1000 - 8000 \\
\hline
Low frequency data weight & 0.0 & 10.0 \\
\hline
\end{tabular}
\end{table}

\begin{table}[htbp] 
\centering
\caption{Parameter settings for Cases B.2--B.4}
\label{tab:parameters_b234}
\begin{tabular}{|l|c|c|c|}
\hline
\textbf{Parameter} & \textbf{Case B.2} & \textbf{Case B.3} & \textbf{Case B.4} \\
\hline
Number of domain points & 10000 & 10000 & 2500 \\
\hline
Number of boundary points & 400 & 400 & 400 \\
\hline
Learning rate & \(1\times 10^{-3}\) & \(5\times 10^{-4}\) & \(1\times 10^{-3}\) \\
\hline
Adam iterations & 5000 & 5000 & 5000 \\
\hline
L-BFGS iterations & 15000 & 5000 & 3000 \\
\hline
Optimizer decay & 1000, 0.9 & 1000, 0.9 & 1000, 0.9 \\
\hline
Network width & 50 & 50 & 60 \\
\hline
Network depth & 6 & 5 & 6 \\
\hline
\end{tabular}
\end{table}

\begin{table}[htbp]
\centering
\caption{Reweighting configuration for Cases B.2--B.4}
\label{tab:reweighting_b234}
\begin{tabular}{|l|c|c|c|}
\hline
\textbf{Parameter} & \textbf{Case B.2} & \textbf{Case B.3} & \textbf{Case B.4} \\
\hline
Decay epsilon & 1.0 & 0.5 & 0.5 \\
\hline
Number of subregions & [5, 5] & [5, 5] & [5, 5] \\
\hline
Reweight every & 1000 & 500 & 500 \\
\hline
Reweight causal steps & 1000 - 9000 & 1000 - 5000 & 1000 - 3000 \\
\hline
Reweight adaptive steps & 10000 - 15000 & 5000 - 8000 & 6000 - 8000 \\
\hline
Low frequency data weight & 10.0 & 10.0 & 10.0 \\
\hline
Reweight adaptive scale & 3.0 & 3.0 & 3.0 \\
\hline
Grad norms scale $\lambda_g$ & 0.5 & 2.0 & 0.5 \\
\hline
\end{tabular}
\end{table}

%%%% Bibliography  %%%%%%%%%%

% \bibliographystyle{plain}
\bibliographystyle{unsrt}
\bibliography{references}

\end{document}